\pdfoutput=1

\documentclass[11pt]{article}

\usepackage[preprint]{acl}
\usepackage{algorithm}
\usepackage{algorithmic}
\usepackage[algo2e,ruled,vlined]{algorithm2e}
\usepackage{times}
\usepackage{latexsym}
\usepackage{subfigure}
\usepackage{graphicx}
\usepackage{amsmath}
\usepackage{amsfonts} 
\usepackage{tabularray}
\usepackage{booktabs}
\usepackage[T1]{fontenc}

\usepackage[utf8]{inputenc}

\usepackage{microtype}

\definecolor{brightmaroon}{rgb}{0.76, 0.13, 0.28}
\definecolor{green(ncs)}{rgb}{0.0, 0.62, 0.42}
\usepackage{inconsolata}

\title{CofiPara: A Coarse-to-fine Paradigm for Multimodal Sarcasm Target Identification with Large Multimodal Models}

\author{Hongzhan Lin$^{\heartsuit}$\thanks{\; Equal contribution.}, Zixin Chen$^{\clubsuit *}$, Ziyang Luo$^{\heartsuit}$, Mingfei Cheng$^{\diamondsuit}$, Jing Ma$^{\heartsuit}$\thanks{\; Corresponding authors.}, Guang Chen$^{\clubsuit \dagger}$ \\
        $^{\heartsuit}$Hong Kong Baptist University\\
        $^{\clubsuit}$Beijing University of Posts and Telecommunications \\ 
        $^{\diamondsuit}$Singapore Management University\\
        \texttt{\{cshzlin,majing\}@comp.hkbu.edu.hk},
        \texttt{\{mailboxforvicky,chenguang\}@bupt.edu.cn}}

\begin{document}
\maketitle
\begin{abstract}
Social media abounds with multimodal sarcasm, and identifying sarcasm targets is particularly challenging due to the implicit incongruity not directly evident in the text and image modalities. Current methods for Multimodal Sarcasm Target Identification (MSTI) predominantly focus on superficial indicators in an end-to-end manner, overlooking the nuanced understanding of multimodal sarcasm conveyed through both the text and image. This paper proposes a versatile MSTI framework with a coarse-to-fine paradigm, by augmenting sarcasm explainability with reasoning and pre-training knowledge. Inspired by the powerful capacity of Large Multimodal Models (LMMs) on multimodal reasoning, we first engage LMMs to generate competing rationales for coarser-grained pre-training of a small language model on multimodal sarcasm detection. We then propose fine-tuning the model for finer-grained sarcasm target identification. Our framework is thus empowered to adeptly unveil the intricate targets within multimodal sarcasm and mitigate the negative impact posed by potential noise inherently in LMMs. Experimental results demonstrate that our model far outperforms state-of-the-art MSTI methods, and markedly exhibits explainability in deciphering sarcasm as well.
\end{abstract}

\section{Introduction}

Sarcasm, a prevalent form of figurative language, is often used in daily communication to convey irony, typically implying the opposite of its literal meaning~\citep{joshi2017automatic}. As an important component in deciphering sarcasm, automated Sarcasm Target Identification (STI) is crucial for Natural Language Processing (NLP) in customer service~\citep{davidov2010semi}, opinion mining~\citep{riloff2013sarcasm}, and online harassment detection~\citep{yin2009detection}. Although prior research on STI has primarily centered on textual content~\citep{joshi2018sarcasm, parameswaran2019detecting}, the surge in multimodal user-generated content has propelled the field of multimodal sarcasm target identification to the forefront of research~\citep{wang2022multimodal}, making it a significant area of study in both NLP applications and multimedia computing. 

The MSTI task is to extract the entities being ridiculed (i.e., sarcasm targets) from both the text and image in multimodal sarcastic content. Previous work~\citep{devlin2019bert, bochkovskiy2020yolov4} attempted to straightforwardly integrate a BERT-based textual encoder and a CNN-based visual encoder for just modeling the sarcasm text and image, respectively. The state-of-the-art approach~\citep{wang2022multimodal} treats MSTI merely as an end-to-end task, primarily focusing on the superficial signals evident in the surface-level text and image. However, a more thorough investigation and understanding of the underlying meanings are essential, particularly in cases where the correlation between image and text is not immediately apparent in multimodal sarcasm~\citep{tian2023dynamic}.


{
\begin{figure}[t]
\subfigure[]{
\begin{minipage}[t]{0.5\linewidth}
\centering
\scalebox{0.75}{\includegraphics[width=5cm]{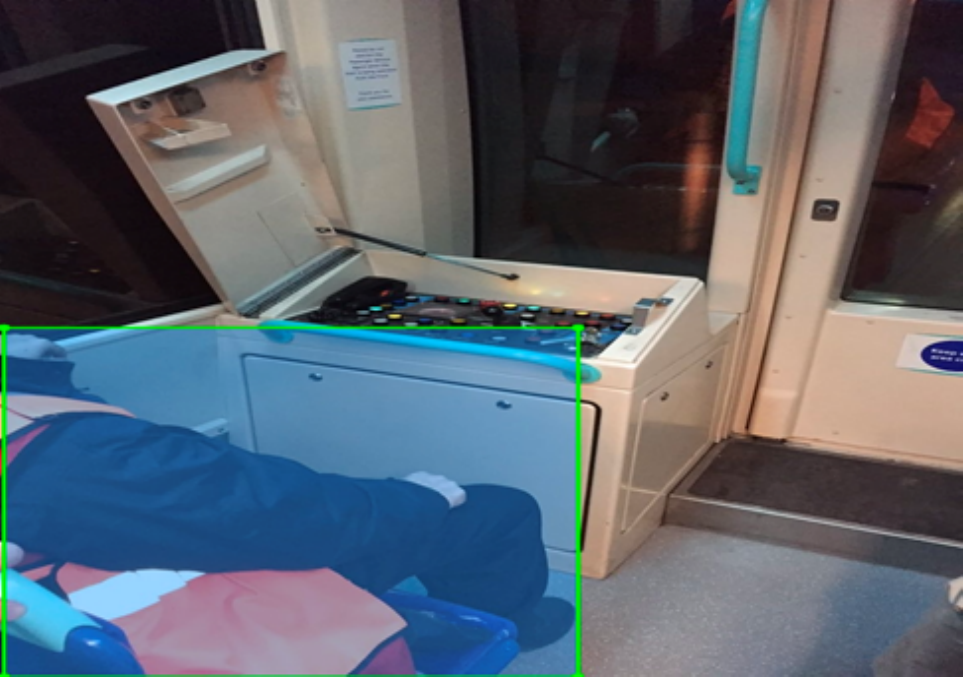}}
\label{fig:sarcasm_1}
\end{minipage}%
}%
\subfigure[]{
\begin{minipage}[t]{0.5\linewidth}
\centering
\scalebox{0.75}{\includegraphics[width=5cm]{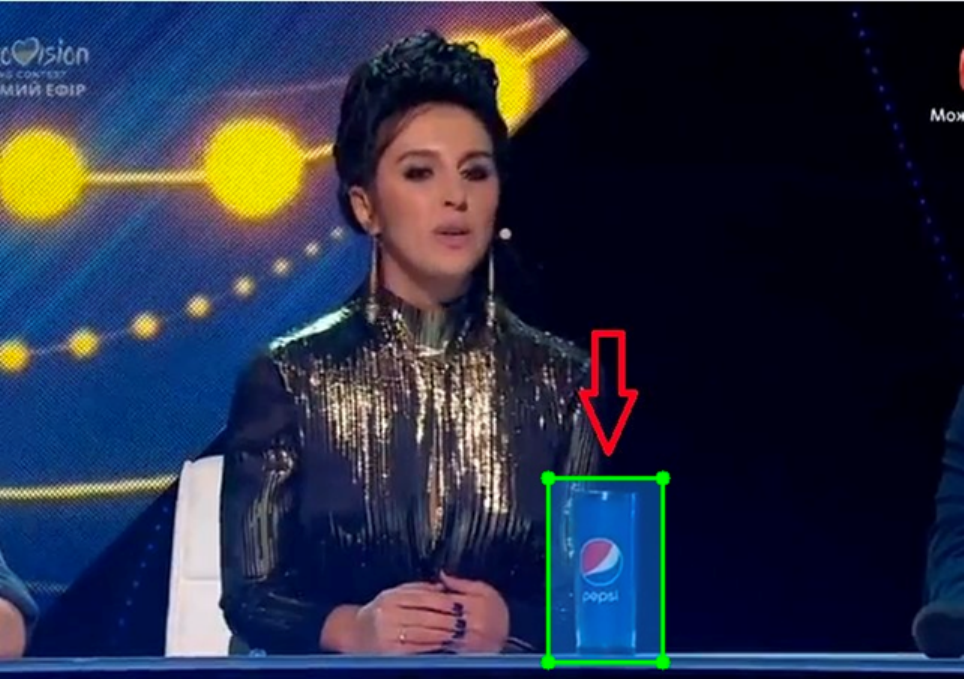}}
\label{fig:sarcasm_2}
\end{minipage}%
}%
\centering
\vspace{-0.4cm}
\caption{Examples of multimodal sarcasm on \textbf{Twitter}: (\textbf{a}) ``\textit{never seen a \#\textcolor{red}{dlr train driver} before. looks like a tough job \#london}''; (\textbf{b}) ``\textit{thank god for no product placement in \#ukraine \#eurovision}''. Boxes in \textcolor{green(ncs)}{green} and words in \textcolor{red}{red} denote the visual and textual targets.}
\label{fig:motivation}
\vspace{-0.4cm}
\end{figure}}

Comprehending and analyzing multimodal sarcasm poses a considerable challenge, because its implicit meaning demands an in-depth understanding and reasoning of commonsense knowledge. For example, as shown in Figure~\ref{fig:sarcasm_1}, a human checker needs reasonable thoughts between visual and textual sarcasm targets, to understand that the man's leisurely sitting posture in front of the control panel creates a sarcastic contrast between the idea of the train driver's job being difficult and the actual scene. Moreover, as the example shows in Figure~\ref{fig:sarcasm_2}, the sarcasm target sometimes does not appear explicitly in the text, which makes it more challenging for conventional models to cognize that the example implies that the presence of the Pepsi bottle is a form of product placement, which is often seen as a marketing tactic. We contend the challenge lies in delivering rich multimodal knowledge that consistently assists in deciphering the concealed semantics within the multimodal nature of sarcasm.
 
In this paper, we adhere to the following two key principles in the knowledge-augmented design of our approach: 1) LMM Reasoning: To grasp the implicit meanings intrinsic in sarcasm, we resort to the extensive prior knowledge embedded within Large Multimodal Models (LMMs)~\citep{liu2023visual, bai2023qwen}. This design philosophy enables complex reasoning, thereby enhancing both the MSTI accuracy and explainability; 2) MSD Pre-training: Previous literature~\citep{joshi2018sarcasm} indicates that MSTI inherently appraises the presence of sarcasm targeting each entity within sarcastic content. Similar to Multimodal Sarcasm Detection (MSD)~\citep{qin2023mmsd2}, which involves determining the sarcasm in multi-modalities at a holistic content level, MSTI actually engages in finer-grained sarcasm detection at the localized entity level. Considering such a close correlation between MSD and MSTI, it is assumed that insights from the coarser-grained MSD are instrumental in discerning sarcasm targets in the finer-grained MSTI. Thus we devise a cohesive framework to operate on the coarse-to-fine training paradigm, aimed at pinpointing nuanced visual and textual targets of multimodal sarcasm for MSTI by benefitting from LMM reasoning and MSD pre-training.

To these ends, we propose a novel framework with a \textbf{\underline{Co}}arse-to-\textbf{\underline{fi}}ne \textbf{\underline{Para}}digm, \textbf{CofiPara}, by leveraging the divergent knowledge extracted from LMMs for multimodal sarcasm target identification. Specifically, we integrate text and image modalities within the coarse-to-fine training paradigm, which consists of two phases: 1) Coarser-grained Sarcasm Detection: Initially, we engage LMMs in critical thinking to generate rationales from both sarcastic and non-sarcastic perspectives. Utilizing these generated sarcasm rationales, we pre-train a smaller model to act as a rationale referee to implicitly extract sarcasm-indicative signals in the competing rationales for sarcasm prediction. This process aligns multimodal features between the sarcasm content and its underlying rationales, alleviating the negative impact of inevitable noise from LMMs through competing rationales; 2) Finer-grained Target Identification: Subsequently, we further fine-tune the smaller model pre-trained in the previous stage for multimodal sarcasm target identification. This phase enhances our model with the multimodal reasoning knowledge, acquired in the pre-training stage and the rationale in sarcastic perspective, to reveal the meanings concealed within the comprehensive multimodal information of sarcasm samples. In this manner, our CofiPara framework could be naturally output as the explanatory basis for deciphering multimodal sarcasm. Extensive experiments conducted on two public sarcasm datasets reveal that our approach far outperforms previous state-of-the-art MSTI methods, and achieves competitive results compared with MSD baselines. The experimental analysis further underscores the enhanced ability to provide superior explainability in the realm of multimodal sarcasm. Our contributions are summarized as follows in three folds:
\begin{itemize}
    \item To the best of our knowledge, we are the first to study multimodal sarcasm from a fresh perspective on explainability in both multimodal targets and natural texts, by exploiting advanced large multimodal models.
    \footnote{Our source code is available at \url{https://github.com/Lbotirx/CofiPara}.}
    \item We propose a universal MSTI framework with the novel coarse-to-fine paradigm that incorporates the multimodal sarcasm target identification and the textual explanation for deciphering the multimodal sarcasm, which enhances sarcasm explainability in conjunction with effective multimodal sarcasm detection.
    \item Extensive experiments confirm that our framework could yield superior performance on multimodal sarcasm target identification, and further provide informative explanations for a better understanding of multimodal sarcasm.
\end{itemize}

\section{Related Work}
\textbf{MSD.}
Sarcasm detection involves discerning sentiment incongruity within a context, traditionally emphasizing text modality~\citep{xiong2019sarcasm, babanejad2020affective}. Multimodal Sarcasm Detection (MSD), enhanced by image integration, has garnered growing research interest~\citep{schifanella2016detecting}. \citet{cai2019multi} introduced a comprehensive MSD dataset, incorporating text, image, and image attributes, alongside a hierarchical fusion model. Subsequently, a range of studies has utilized attention mechanisms to subtly blend features from different modalities~\citep{xu2020reasoning, pan2020modeling, tian2023dynamic}. Another line of recent advancements has seen the introduction of graph-based methods for sarcasm detection~\citep{liang2021multi, liang2022multi, liu2022towards}, which excel in identifying key indicators across modalities. \citet{qin2023mmsd2} revealed spurious cues in the previous MSD dataset~\citep{cai2019multi} and provided an alternative refined dataset version. Existing solutions, however, only focused on performing multimodal sarcasm classification (i.e., predicting if a sample is sarcastic) with limited explanations for its prediction. In this paper, we delve into the explainability of multimodal sarcasm from both multimodal targets and textual rationales, aiming to decipher multimodal sarcasm using more intuitive forms and assisting users in gaining a better understanding.

\textbf{MSTI.}
Recent advancements in sarcasm analysis have seen a significant focus on Sarcasm Target Identification (STI), with notable contributions from researchers. STI aims to pinpoint the subject of mockery in sarcastic texts. \citet{joshi2018sarcasm} introduced the concept of STI and discussed its application in the 2019 ALTA shared task~\citep{molla2019overview}, highlighting evaluation metrics like Exact Match accuracy and F1 score. \citet{patro2019deep} later developed a deep learning model enhanced with socio-linguistic features for target identification, while \citet{parameswaran2019detecting} utilized a combination of classifiers, 
followed by a rule-based method for extracting textual sarcasm targets. Moreover, \citet{wang2022multimodal} pioneered STI in multimodal contexts by integrating sequence labeling with object detection in an end-to-end manner, but only capturing the superficial signals of different modalities in sarcasm. In this work, we regard the MSD task as the predecessor pre-training phase of MSTI, to better derive the prior reasoning knowledge absorbed in the coarser-grained auxiliary task MSD to the finer-grained goal task MSTI.

\textbf{LLMs and LMMs.}
Recently, Large Language Models (LLMs) have demonstrated exceptional versatility across various tasks. Significant advancements by leading tech companies have resulted in highly proficient, though often proprietary, LLMs~\citep{brown2020language, OpenAI2023GPT4TR, chowdhery2022palm, team2023gemini}. Meanwhile, the NLP community has seen the rise of open-source LLMs, with publicly shared model weights~\citep{black2022gpt, zeng2022glm, touvron2023llama, touvron2023llama2, wizardlm, wizardcoder}. More recently, LLMs have also been developed to adapt in processing both textual and visual data, marking a significant advancement. Recent research has focused on constructing versatile multimodal datasets~\citep{yang2023dawn} from platforms like GPT-4 and GPT-4V~\citep{OpenAI2023GPT4TR}, fine-tuning open-source LMMs such as LLaVA~\citep{liu2023visual}, Qwen-VL~\citep{bai2023qwen}, and other innovative projects~\citep{Dai2023InstructBLIPTG, wang2023cogvlm}. These LMMs have shown excellent emergent abilities in multimodal tasks. In this work, we foster divergent thinking in LMMs by employing potential sarcasm labels as prompts, which promotes a coarse-to-fine strategy for fine-tuning smaller Language Models (LMs). Combined with MSTI, this design philosophy enhances the sarcasm understanding within the universal framework, steering it towards greater sarcasm explainability.

\section{Our Approach}
\begin{figure*}
    \centering
    \includegraphics[width=\linewidth]{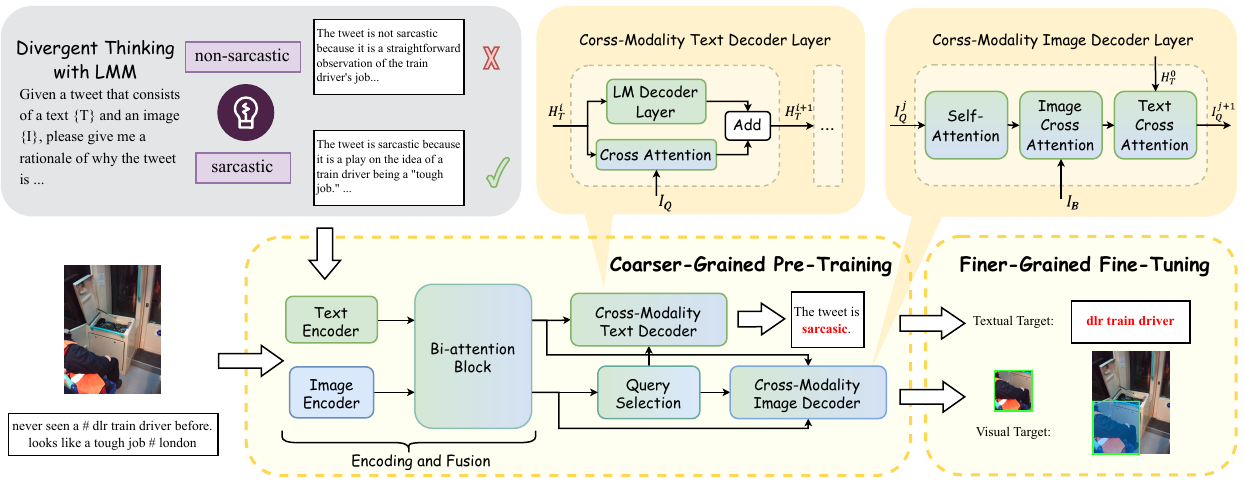}
    \vspace{-0.7cm}
    \caption{An overview of our framework, CofiPara, for multimodal sarcasm target identification.} 
    \label{fig:model}
    \vspace{-0.5cm}
\end{figure*}

\textbf{Problem Statement.}
We define a multimodal sample as $M=\{I, T\}$, which consists of an image $I$ and a text $T$. In the context of the coarser-grained MSD task, the label $y$ of the sample falls into either of the categories: {\texttt{sarcastic}} or {\texttt{non-sarcastic}}.  As for the finer-grained MSTI task, the label $y$ is a tuple consisting of a textual sarcasm target $y_{text}$, and a visual bounding box $y_{img}$, where the model is tasked with pinpointing the sarcasm entity targeted within the provided text and image modalities of the \textit{sarcastic} sample. In this paper, we focus on improving the finer-grained MSTI task by leveraging insights from the coarser-grained MSD task.

Closed to the MSD task, which establishes the presence of sarcasm in holistic semantics at the coarser level, the MSTI task inherently detects sarcasm targeting each entity of the multimodal sarcastic content to explicitly identify the specific sarcasm targets at the finer level~\citep{joshi2018sarcasm}. This work is designed mainly to integrate MSD and MSTI into a versatile framework with the coarse-to-fine training paradigm, which utilizes MSD as the predecessor foundational stage to facilitate the subsequent MSTI process, through incorporating rationales generated from LMMs.
The overview of our model is illustrated in Figure \ref{fig:model}, which consists of: 1) Divergent Thinking with LMM ($\S$\ref{LMM}), 2) Coaser-Grained Pre-Training ($\S$\ref{MSD}), and 3) Finer-Grained Fine-Tuning ($\S$\ref{MSTI}).

\subsection{Divergent Thinking with LMM}\label{LMM}

Generally, LLMs can generate reasonable thoughts~\citep{wei2022chain} to unveil the underlying meaning of the sarcasm. The rationales from LLMs usually express perspectives grounded with commonsense or related to certain social scenarios~\citep{huang2023chatgpt, lin2024explainable}, which can be used as extensive prior knowledge for smaller downstream models to facilitate decision-making.
Although LLMs have shown emergent abilities in reasoning and interpreting, they still suffer from preconception bias and may generate uncredible content or even make false assertions~\citep{hu2023bad, lin2023beneath}. Therefore, the downstream decision-maker model would have to be robust enough to alleviate the negative impact imposed by the input noisy LLM-generated rationales. In pursuit of this, we resort to the inspiration of divergent multimodal thinking with vision LLMs, i.e., LMMs, fostering our model to explore a more reliable reasoning pathway from the conflicting and noisy insights provided by LMMs. 

Given an input ${M} = \{I,T\}$, we prompt LMMs to generate a pair of competing rationales based on the text $T$, the image $I$, and the potential sarcasm labels ${*} \in \{\texttt{sarcastic}, \texttt{non-sarcastic}\}$, by using a prompt template $p$ we curated in advance. To exploit LMMs' divergent reasoning ability, we construct each sample with different potential sarcasm labels, respectively. Specifically, we design the prompt template $p$ as follows:

\textit{``Given a tweet that consists of a text and an image, please give me a rationale of why the tweet is $\{*\}$. \\
tweet text: $\{T\}$ \\
tweet image: $\{I\}$''
}

Note that the potential labels $*$ are just used to formalize two opposite standpoints for multimodal reasoning regardless of the ground-truth label. Then we can derive the competing rationales $r_{pos}$ or $r_{neg}$ from LMMs to support the \texttt{sarcastic} or \texttt{non-sarcastic} positions.
By introducing adversarial labels, we encourage LMMs to adopt diverse perspectives, thereby providing a range of background knowledge enriched with deliberate noise. Because the rationale for the false class ideally contains more useless information than the other one for the ground truth. The contextual subtleties of sarcasm that are pivotal to rival candidate sarcasm categories, can be thus more effectively highlighted and contrasted. This allows the rest of the model to achieve the logical reasoning of the true sarcastic intent by considering it from diverse perspectives, while moderating vulnerability to the potential noise in the LMM-generated rationales.

\subsection{Coarser-Grained Pre-Training}\label{MSD}

Given the close correlation between the coarser-grained multimodal sarcasm detection with the finer-grained multimodal sarcasm target identification, we advocate for initial pre-training in multimodal sarcasm detection, allowing the model to grasp the essence of sarcasm preliminarily. This foundational understanding could set the stage for a more nuanced and detailed identification of sarcasm targets in subsequent fine-tuning.

\noindent\textbf{Encoding and Fusion.} 
For an input sample ${M} = \{I,\hat{T}\}$ packed with the generated competing rationales, where $\hat{T} = \{T, r_{pos}, r_{neg}\}$ is the input text, we first extract textual and visual features as:
{
\setlength{\abovedisplayskip}{0.1cm}
\setlength{\belowdisplayskip}{0.1cm}
\begin{equation}
H_T =\mathsf{E}_T(\hat{T}), 
~~
I_E =\mathsf{E}_I(I),
\end{equation}}
where $H_{T} \in \mathbb{R}^{m \times d}$ is the token embedding output by the text encoder $\mathsf{E}_T(\cdot)$ implemented by Transformer Encoder~\citep{raffel2020exploring}, $m$ is the input token length and $d$ is the dimension of hidden states. $\mathsf{E}_I(\cdot)$ denotes the image encoder based on Vision Transformer~\citep{liu2021swin}, used to fetch the patch-level features of the image with $n$ patches, which are projected into the visual features $I_E \in \mathbb{R}^{n \times d}$. Since both the text and image encoders are designed as Transformer-based, the embeddings shaped by the isomorphic encoding structure can enhance consecutive multimodal fusion during encoding~\citep{liu2023grounding}.
Then, to align the semantics in the text and image, we adopt a bi-directional attention module based on the cross-attention mechanism. Taking the text-to-image cross-attention ${CrossAttn}(H_T,I_E)$ as an example, we define the query, key and value as $\{Q_{{T}},  K_{{I}}, V_{{I}}\} = \{H_T W_Q, I_E W_K, I_E W_V\}$, where $\{W_{Q}, W_{K}, W_{V}\} \in \mathbb{R}^{d \times d_k}$ are trainable weights. Then the calculation is as follows:
{
\setlength{\abovedisplayskip}{0.1cm}
\setlength{\belowdisplayskip}{0.1cm}
\begin{equation}
    \begin{aligned}
H_{T}^0  & =\operatorname{softmax}\left(\frac{Q_T K_I^{\top}}{\sqrt{d_k}}\right) V_I,
\end{aligned}
\label{c_attn}
\end{equation}}
With Equation~\ref{c_attn}, similarly, we can calculate the image-to-text cross-attention. Combing the two cross-attention modules together, we fuse the multimodal features during encoding as follows:
{
\setlength{\abovedisplayskip}{0.1cm}
\setlength{\belowdisplayskip}{0.1cm}
\begin{equation}
    \begin{aligned}
H_{T}^0 &= {CrossAttn}(H_T, I_E), \\
I_B &= {CrossAttn}(I_E,H_T),
\end{aligned}
\end{equation}} where $H_{T}^0, I_B$ are the attended textual and visual features, respectively. 
%
To optimize the information integration of multimodal sarcasm with competing rationales, we further develop a query selection mechanism to prioritize image region features that exhibit higher correlations with the input text. This yields:
{
\setlength{\abovedisplayskip}{0.1cm}
\setlength{\belowdisplayskip}{0.1cm}
\begin{equation}
\arg \max_n \left(\max_m \left(I_B {H_{T}^0}^{\top} \right)\right),
\end{equation}}
with which we obtain the index of the topmost relevant local visual features, queried by the textual features ${H_{T}^0}$ from the global visual features $I_B$. We name the selected local visual features as $I_Q$.

\noindent\textbf{Cross-Modality Text Decoding.}  Based on the attended textual features $H_{T}^0$ and query-selected local visual features $I_Q$, we then devise a multimodal decoding strategy with textual outputs to infer sarcasm for MSD. Specifically, during decoding, we only exploit a text-to-image cross-attention module to attain the textual features attended with the visual ones:
{
\setlength{\abovedisplayskip}{0.1cm}
\setlength{\belowdisplayskip}{0.1cm}
\begin{equation}
    \begin{aligned}
{H_T^i}_{attn} &= {CrossAttn}(H_{T}^i, I_Q),
\end{aligned}
\end{equation}}
where $H_T^i$ is the textual feature input of the $i^{th}$ LM decoder layer, and ${H_T^i}_{attn}$ is the attended textual feature output of the cross-attention ${CrossAttn}$. Then by adding the attended features ${H_T^i}_{attn}$ to the output of the $i^{th}$ LM decoder $LM_{dec}^i$, the fused intermediate features $H_T^{i+1}$ fed into the next LM decoder layer are:
{
\setlength{\abovedisplayskip}{0.1cm}
\setlength{\belowdisplayskip}{0.1cm}
\begin{equation}
\begin{aligned}
H_T^{i+1} & = LM_{dec}^i(H_T^i) + {H_T^i}_{attn}.
\end{aligned}
\end{equation}}
After $L$ layers of cross-modality LM decoder, we have the final textual representations $H_T^L$, further decoded as the text output to clearly express whether the sample is sarcastic. Finally, we train the model $f$ by minimizing the following loss:
{
\setlength{\abovedisplayskip}{0.1cm}
\setlength{\belowdisplayskip}{0.1cm}
\begin{equation}
\mathcal{L}_{text} = CE(f(I,\hat{T}),y),
\label{eq6}
\end{equation}}
where $CE(\cdot)$ denotes the cross-entropy loss between the generated label token and ground truth label $y$ for MSD. During the coarser-grained pre-training, the model is trained to distill the essence and discard irrelevant elements from the divergent thinking of LMMs about sarcasm. Such a process could fortify our model's resilience in the subsequent fine-tuning stage for MSTI, ensuring robustness against the potential inaccuracies stemming from LMMs, leading to a more independent and refined thought of the LMM-generated rationales.

\subsection{Finer-Grained Fine-Tuning} 
\label{MSTI}
After the coarser-grained pre-training stage, our model could be resilient against the potential variation and bias in LMMs through the competing rationales, to first comprehend what constitutes sarcasm. As the goal of our approach is to identify both the textual and visual sarcasm targets for further deciphering sarcasm, we conduct the finer-grained fine-tuning stage for MSTI, which shares the same model architecture, parameters of the multimodal encoding and text decoding procedures as $\S$\ref{MSD} but differs in the text decoding output and an additional image decoding procedure.

Different from the pre-training stage in $\S$\ref{MSD}, the sample $M$ in MSTI is set as \textit{sarcastic} prior due to the nature of this specific task~\citep{wang2022multimodal}.
Thus the input text for a given \textit{sarcastic} sample $M$ is formed as $\hat{T} = \{T, r_{pos}\}$ in this stage, where we only provide the text $T$ and the LMM-generated rationale $r_{pos}$ that explains why $M$ is sarcastic. For the cross-modality text decoding, we generate the predicted textual sarcasm targets that are entities in the text $T$. Then the textual target loss $\hat{\mathcal{L}}_{text}$ can be computed akin to that outlined in Equation~\ref{eq6}.   

\noindent\textbf{Cross-Modality Image Decoding.} 
For visual object detection, we use a cross-modality image decoder to discern the visual sarcasm target, where the textual features $H_T^0$ and the global visual features $I_B$ are used to attend to the local visual features $I_Q$ in each Transformer decoder layer:
{
\setlength{\abovedisplayskip}{0.1cm}
\setlength{\belowdisplayskip}{0.1cm}
\begin{equation}
\begin{aligned}
& I_Q^{j^{\prime}}={SelfAttn}\left(I_Q^j\right), \\
& I_Q^{j^{\prime \prime}}={CrossAttn}\left(I_Q^{j^{\prime}}, I_B\right), \\
& I_Q^{j+1}={CrossAttn}\left(I_Q^{j^{\prime \prime}}, H_T^0\right),
\end{aligned}
\end{equation}}
where ${SelfAttn}(\cdot)$ denotes self-attention, and $I_Q^j$ is the input of the $j^{th}$ Transformer decoder layer. After $K$ layers of the image decoder, we have the final visual features $I_Q^K$. Afterwards, we decode $I_Q^K$ as the image output consisting of a bounding box output and its confidence score. Following previous object detection work~\citep{zhang2022dino}, we use the L1 loss $\mathcal{L}_{l1}$ and the GIOU~\citep{rezatofighi2019generalized} loss $\mathcal{L}_{giou}$ for bounding box regressions, and the cross-entropy classification loss $\mathcal{L}_{cls}$ for confidence scores as the joint optimization objective:
{
\setlength{\abovedisplayskip}{0.1cm}
\setlength{\belowdisplayskip}{0.1cm}
\begin{equation}
\mathcal{L}_{img} = \alpha  \mathcal{L}_{l1} + \beta  \mathcal{L}_{giou} +\gamma  \mathcal{L}_{cls},
\label{eq_imgloss}
\end{equation}}
where $\alpha, \beta$ and $\gamma$ are the hyper-parameters to scale the losses, $\mathcal{L}_{img}$ is the visual target loss. Finally, the overall training loss $\mathcal{L}$ for this stage is:
{
\setlength{\abovedisplayskip}{0.1cm}
\setlength{\belowdisplayskip}{0.1cm}
\begin{equation}\mathcal{L} = \mathcal{L}_{img} + \hat{\mathcal{L}}_{text}.
\end{equation}}

\paragraph{Model Training.}
We implement model training following a coarse-to-fine paradigm: 1) Pre-training on the coarser-grained MSD task by minimizing $\mathcal{L}_{text}$, and 2) Fine-tuning on the finer-grained MSTI task by minimizing $\mathcal{L}$, where the auxiliary task MSD is the predecessor training phase of the goal task MSTI. To this end, we unify the classification task for MSD and the sequence tagging task for textual target identification in MSTI into a text generation task. Note that for model testing on MSD, we use the model parameters obtained after the coarser-grained pre-training; in terms of the goal task MSTI, we directly input the test sarcastic sample into our finer-grained fine-tuned model to identify multimodal sarcasm targets. 

\section{Experiments}
\subsection{Experimental Setup}
\textbf{Datasets.} Our experiments are conducted based on two publicly available multimodal sarcasm datasets for evaluation: MMSD2.0~\citep{qin2023mmsd2} and MSTI~\citep{wang2022multimodal}. Specifically, MMSD2.0 is a correction version of the raw MMSD dataset~\citep{cai2019multi}, by removing the spurious cues and fixing unreasonable annotation.
In the coarser-grained pre-training stage, we utilized the large-scale MMSD2.0 dataset to pre-train our model for multimodal sarcasm detection. For the MSTI dataset, as the visual sarcasm target identification in the original MSTI data is mixed with optical character recognition, the reproduction of the code and data released by MSTI\footnote{\url{https://github.com/wjq-learning/MSTI}} has a big gap compared with the object detection results reported in their paper~\citep{wang2022multimodal}. Thus we introduce a refined version, i.e., MSTI2.0, to address the low-quality issue of the raw MSTI data by removing the visual target labels in images of only characters and converting them into the textual sarcasm target labels. Afterwards, MSTI2.0 is employed to fine-tune and evaluate the model in the finer-grained fine-tuning stage of our framework. 

\begin{table}[] \small
\centering
\resizebox{0.95\linewidth}{!}{\begin{tabular}{lcccc}
\toprule
\multicolumn{1}{c}{}       & \multicolumn{1}{l}{Acc.} & P     & R     & F1    \\ \hline
Att-BERT                   & 80.03                    & 76.28 & 77.82 & 77.04 \\
CMGCN                      & 79.83                    & 75.82 & 78.01 & 76.90 \\
HKE                      & 76.50                    & 73.48 & 71.07 & 72.25 \\
DynRT                      & 72.06                    & 71.79 & 72.18 & 71.98 \\
Multi-view CLIP           & 84.31                    & 79.66 & 85.34 & 82.40 \\ \hline
CofiPara-\textsc{Msd}               & \textbf{85.70}                    & \textbf{85.96} & \textbf{85.55} & \textbf{85.89} \\
\toprule 
\end{tabular}}
\vspace{-0.3cm}
\caption{Multimodal sarcasm detection results.}
\label{msd_res}
\vspace{-0.4cm}
\end{table}

\noindent\textbf{Baselines.} We compare our model with the following multimodal baselines for multimodal sarcasm detection, which is the auxiliary task: 
1) \textsf{Att-BERT}~\citep{pan2020modeling}; 2) \textsf{CMGCN}~\citep{liang2022multi}; 3) \textsf{HKE}~\citep{liu2022towards}; 4) \textsf{DynRT-Net}~\citep{tian2023dynamic}; 5) \textsf{Multi-view CLIP}~\citep{qin2023mmsd2}. We adopt Accuracy, F1 score, Precision, and Recall to evaluate the MSD performance.

To evaluate our model in multimodal sarcasm target identification that is our goal task, we compare the following state-of-the-art MSTI systems: 1) \textsf{BERT-Base}~\citep{devlin2019bert}; 2) \textsf{BERT-Large}; 3) \textsf{Mask R-CNN}~\citep{he2017mask}; 4) \textsf{YOLOv8}~\citep{terven2023comprehensive}; 5) \textsf{OWL-ViT}~\citep{minderer2022simple}; 6) \textsf{Grounding DINO}~\citep{liu2023grounding}; 7) \textsf{MSTI-RB}~\citep{wang2022multimodal}; 8) \textsf{MSTI-VB}; 9) \textsf{MSTI-CB}; 10) \textsf{MSTI-CL}. We use Exact Match (EM)~\citep{joshi2018sarcasm} and F1 score~\citep{molla2019overview} as evaluation metrics of textual sarcasm target identification; and Average Precision (AP)~\citep{lin2014microsoft}, i.e., the COCO-style AP, AP50, and AP75, as the metrics for visual sarcasm target identification. 

The data statistics, construction details of MSTI2.0, baseline descriptions and model implementation are detailed in the Appendix.

\begin{table*}[]
\resizebox{\textwidth}{!}{
\begin{tabular}{lcccccccccc}
\toprule
\multicolumn{1}{c}{}              & \multicolumn{5}{c}{Dev}               & \multicolumn{5}{c}{Test}              \\ \cmidrule(lr){2-6} \cmidrule(l){7-11}
                                  & EM    & F1    & AP    & AP50  & AP75  & EM    & F1    & AP    & AP50  & AP75  \\ \hline
BERT-Base                         & 26.82 & 45.23 & /     & /     & /     & 26.01 & 46.64 & /     & /     & /     \\
BERT-Large                        & 29.29 & 46.42 & /     & /     & /     & 27.89 & 46.93 & /     & /     & /     \\ \hline
Mask R-CNN                        & /     & /     & 06.90 & 13.30 & 05.70 & /     & /     & 07.60 & 14.30 & 07.30 \\
YOLOv8                            & /     & /     & 06.58 & 12.81 & 06.13 & /     & /     & 10.49 & 17.57 & 11.18 \\ \hline
OWL-ViT                           & 14.80 & 01.20 & 03.36 & 13.75 & 00.17 & 18.40 & 01.64 & 03.32 & 14.47 & 00.91 \\
Grounding DINO                    & 18.29 & 01.60 & 11.15 & 19.77 & 10.37 & 15.22 & 00.59 & 10.92 & 17.26 & 11.30  \\
MSTI-RB (ResNet+BERT-Base)        & 27.09 & 47.28 & 01.82 & 06.71 & 00.14 & 28.84 & 47.05 & 02.11 & 07.80 & 00.30 \\
MSTI-VB (VGG19+BERT-Base)         & 28.19 & 45.74 & 02.03 & 07.43 & 00.27 & 29.51 & 49.02 & 02.57 & 08.92 & 00.24 \\
MSTI-CB (CSPDarkNet53+BERT-Base)  & 27.62 & 48.00 & 03.78 & 13.68 & 00.40 & 27.89 & 48.39 & 03.80 & 13.06 & 01.03 \\
MSTI-CL (CSPDarkNet53+BERT-Large) & 28.18 & 48.32 & 02.64 & 09.56 & 00.86 & 28.70 & 49.78 & 02.80 & 11.02 & 00.91 \\ \hline
CofiPara-\textsc{Msti}                    & \textbf{31.96} & \textbf{49.53} & \textbf{15.38} & \textbf{34.29} & \textbf{15.57} & \textbf{32.26} & \textbf{50.27} & \textbf{13.79} & \textbf{32.49} & \textbf{12.01} \\ \toprule
\end{tabular}}
\vspace{-0.3cm}
\caption{Multimodal sarcasm target identification results.}
\label{msti_res}
\vspace{-0.4cm}
\end{table*}

\subsection{Main Results}

\textbf{Sarcasm Detection Performance.} Table \ref{msd_res} illustrates the performance (\%) of our proposed method versus all the compared representative multimodal baselines on the auxiliary task MSD. From these results, we have the following observations: 1) Compared to graph-based methods such as CMGCN and HKE and routing-based DynRT, Att-BERT that relies on semantic understanding has better performance, indicating that this task requires models to capture deep semantic information rather than superficial attributes. 2) Multi-view CLIP shows an overall advantage in its ability to align textual and visual features, and the isomorphic structures of text and image encoder also contribute to its superiority. 3) Our proposed CofiPara-\textsc{Msd} surpasses the leading baseline by 1.39\% and 3.49\% in accuracy and F1 score, additionally demonstrating a more balanced performance in terms of recall and precision, despite not primarily targeting the MSD task. The distinctive advantage of our model lies in the fact that while all the baselines solely focus on recognition, our model is equipped with rationales from divergent thinking with LMMs, which empowers our model to effectively uncover sarcastic content by adeptly leveraging the interplay between seemingly unrelated textual and visual elements within sarcasm.

\noindent\textbf{Target Identification Performance.}
Table \ref{msti_res} shows the performance (\%) of our method versus unimodal and multimodal baselines on the goal task MSTI. It can be observed that: 1) The unimodal methods, like text-modality models in the first group and image-modality models in the second group, fall short in simultaneously identifying both visual and textual sarcasm targets compared to the multimodal methods in the third group. 2) The textual target identification performance of visual grounding models (i.e., OWL-ViT and Grounding DINO), is hindered by the discrepancy between the MSTI task and their original pre-training objectives. Additionally, the lack of a consistent one-to-one correspondence between textual and visual targets in MSTI samples further contributes to their suboptimal performance. 3) Our method drastically excels in EM and AP50 compared to baselines, especially in visual target identification. We observe that CofiPara-\textsc{Msti} improves textual target identification performance by 3.26\% on average EM score compared to MSTI-VB, suggesting that our model is more precise in discerning sarcasm targets in the text modality of multimodal contents. On the other hand, our model exhibits a substantial superiority in visual target identification performance, especially on the AP50 metric, for an average improvement of 14.88\% over the best visual performed baseline, indicating that our model can capture the correct visual targets within the image modality that contain sarcastic meanings, while baseline models perform poorly by simply identifying object rather than sarcasm targets, which further implies that our model displays a better understanding of multimodal sarcasm.

\subsection{Ablation Study of Target Identification}

\begin{table}[]
\resizebox{\linewidth}{!}{\begin{tabular}{lccccc}
\toprule
                               & EM    & F1    & AP    & AP50  & AP75  \\ \hline
CofiPara-\textsc{Msti}                  & 32.26 & 50.27 & 13.79 & 32.49 & 12.01 \\ \hline
w/o MSD               & 30.24 & 49.61 & 13.72 & 30.39 & 12.15 \\
w/o LMM                 & 30.91 & 48.32 & 07.50 & 19.36 & 04.22 \\
w/o MSD\&LMM & 30.10 & 50.72 & 06.34 & 17.21 & 04.61 \\ \toprule
\end{tabular}}
\vspace{-0.3cm}
\caption{Ablation results on MSTI2.0 test set.}
\label{abla_msti_test}
\vspace{-0.5cm}
\end{table}

As MSTI is our goal task, we conduct ablative studies on MSTI2.0 test data with the following variants: 1) \textit{w/o MSD}: Simply train our model on the MSTI task without knowledge from pre-training on MSD.
2) \textit{w/o LMM}: Use model parameters initialized by pre-training on the MSD task, and fine-tune directly on the MSTI task without knowledge from LMMs. 3) \textit{w/o MSD\&LMM}: Train our model directly on the MSTI task without any knowledge of LMM reasoning and MSD pre-training.

As demonstrated in Table \ref{abla_msti_test}, our model shows different degrees of performance degradation when MSD pre-training or LMM reasoning knowledge is ablated, indicating the effectiveness of our proposed method. Specifically, visual target identification performances show significant degradations by 2.10\% and 13.13\% on AP50 for \textit{w/o MSD} and \textit{w/o LMM} settings, respectively. This indicates that both LMM reasoning and MSD pre-training are helpful in identifying sarcasm targets, and that external LMM knowledge has a relatively larger impact on visual performance. We also notice that, the \textit{w/o LMM} setting has relatively mild improvement over \textit{w/o MSD\&LMM}. This can be attributed to the fact that although the MSD pre-training itself may not necessarily significantly enhance model performance on the MSTI task with a large margin, it could help our model learn to implicitly ignore useless expressions and extract informative signals in the rationales from LMMs, highlighting its synergistic complementary with the LMM knowledge.

\subsection{Case Study of Explainability}
\label{case_study_sec}

\begin{figure}
    \centering
    \includegraphics[width=\linewidth, scale=1.00]{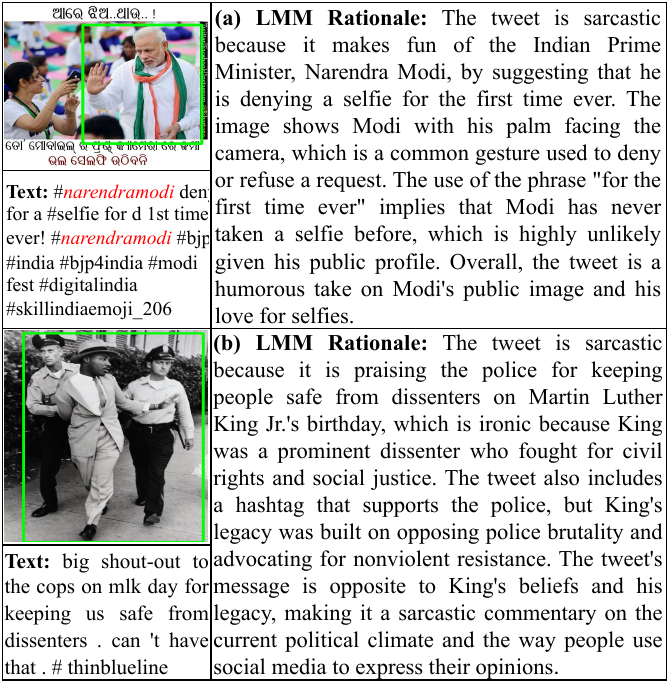}
    \vspace{-0.6cm}
    \caption{Examples of correctly identified samples.} 
    \label{case_study}
    \vspace{-0.5cm}
\end{figure}

To better understand the mechanism of how LMM-generated rationales facilitate sarcasm target identification, we conduct a case study on the correctly predicted samples for better sarcasm explainability, as shown in Figure \ref{case_study}, where visual sarcasm targets are annotated by green rectangles and textual sarcasm targets are highlighted in red italics.

In these examples, we observe that: 1) rationales generated by LMM help promote the connections between two modalities. As shown in Figure \ref{case_study}(a), we notice that in the generated rationale, the image is depicted as a photo of Narendra Modi, which is then linked to the man who makes a refusing gesture in the picture. By introducing the connection between the word ``narendramodi'' and the man in the image, the target can be more easily recognized by our model; 2) on the other hand, rationales can complement background messages that are not given in the original texts and images, including both common sense and political knowledge. For example, in Figure \ref{case_study}(a), the rationale first recognizes the man as Prime Minister of India, and then offers a correction ``for the first time ever'' to the non-standard abbreviation of ``for d 1st time ever'', which is further explained as an expression of sarcastic tone towards Modi. Similarly, in Figure \ref{case_study}(b), LMM interprets ``mlk'' as Martin Luther King Jr. Day, the day to memorize dissenters who fought for civil rights, while the fact that police are arresting the dissenter in the image is in conflict with the context that expresses thanks to police, as well as the hashtag \textit{\#thinblueline}. The sarcasm target in the image is explained as the unjust political situation for people who fight for human rights, which is depicted in the image but outside the text. In this way, the rich but implicit correlations between the sarcasm text and image could be explained in visualized targets and readable snippets, which are also potentially valuable for aiding human checkers in verifying the sarcasm. We provide error analysis and explainability evaluation in the Appendix.

\section{Conclusion and Future Work}
In this paper, we proposed a novel coarse-to-fine paradigm to decode the implicit meanings hidden beneath the surface of texts and images in multimodal sarcasm for MSTI, by leveraging rich prior knowledge from LMM reasoning and MSD pre-training. We first inspired divergent thinking with LMMs to derive competing rationales for coarser-grained pre-training of a small language model on MSD. Then we conducted finer-grained fine-tuning of the model on MSTI. Comprehensive experiments and analyses confirm the advantages of our framework. Future efforts aim to enhance our research by focusing on explicitly extracting useful information from generated rationales, to further relieve the inherent bias and variation in LMMs.

\section*{Limitations}
There are multiple ways to further improve this work:
\begin{itemize}
\item Overall, this work's explainability primarily hinges on the sample's multi-modalities to justify its sarcasm through the visualized targets and readable rationales. Nevertheless, it does not delve into the more profound aspect of explainability concerning the internal workings of neural models. Future efforts will aim to enhance our research by focusing on improving the interpretability of the model's architecture.
\item Generally, the distribution drift in datasets over time is a potential limitation for almost all data-driven tasks, especially for the multimodal sarcasm on social media. However, one of the contributions of this work is proposing a novel paradigm to leverage commonsense reasoning knowledge in LMMs and pre-training insights from MSD for the MSTI task. The proposed framework is general enough, which should still work with newly released stronger LMMs or new sarcasm data appearing on social media. For example, in the future, we could publish a plug-and-play interface to incorporate a broader range of LMMs into our framework.
\item Although sarcasm is defined much differently with hatefulness or offensiveness in previous literature~\cite{xiong2019sarcasm}, in the future, we would try to incorporate more of the related social media and sentiment analysis datasets beyond our task to further broaden the boundaries of this research, such as multimodal target identification of hatefulness, offensiveness, sexism, or cyberbullying, etc.
\item This work mainly focuses on the coarse-to-fine paradigm incorporating competing rationales from LMMs, to implicitly alleviate the negative impact posed by the potential noise in LMM-generated rationales. We would further update our paradigm to explicitly distill and display useful information from LMMs and avoid several common deficiencies of existing language models, including hallucination and limited generalization, as much as possible.
\end{itemize}

\section*{Ethics Statement}
All data of MSTI2.0 come from the original MSTI dataset~\citep{wang2022multimodal}, which is an open-source dataset available for academic research. Our re-annotation and rectification process is designed to be carried out automatically.  All the data in this work only include text and image modalities and do not contain any user information on social media.

\section*{Acknowledgements}
This work is partially supported by National Natural Science Foundation of China Young Scientists Fund (No. 62206233) and Hong Kong RGC ECS (No. 22200722).
\bibliography{custom}

\begin{thebibliography}{65}
\expandafter\ifx\csname natexlab\endcsname\relax\def\natexlab#1{#1}\fi

\bibitem[{Babanejad et~al.(2020)Babanejad, Davoudi, An, and Papagelis}]{babanejad2020affective}
Nastaran Babanejad, Heidar Davoudi, Aijun An, and Manos Papagelis. 2020.
\newblock Affective and contextual embedding for sarcasm detection.
\newblock In \emph{Proceedings of the 28th international conference on computational linguistics}, pages 225--243.

\bibitem[{Bai et~al.(2023)Bai, Bai, Yang, Wang, Tan, Wang, Lin, Zhou, and Zhou}]{bai2023qwen}
Jinze Bai, Shuai Bai, Shusheng Yang, Shijie Wang, Sinan Tan, Peng Wang, Junyang Lin, Chang Zhou, and Jingren Zhou. 2023.
\newblock Qwen-vl: A frontier large vision-language model with versatile abilities.
\newblock \emph{arXiv preprint arXiv:2308.12966}.

\bibitem[{Black et~al.(2022)Black, Biderman, Hallahan, Anthony, Gao, Golding, He, Leahy, McDonell, Phang et~al.}]{black2022gpt}
Sidney Black, Stella Biderman, Eric Hallahan, Quentin Anthony, Leo Gao, Laurence Golding, Horace He, Connor Leahy, Kyle McDonell, Jason Phang, et~al. 2022.
\newblock Gpt-neox-20b: An open-source autoregressive language model.
\newblock In \emph{Proceedings of BigScience Episode\# 5--Workshop on Challenges \& Perspectives in Creating Large Language Models}, pages 95--136.

\bibitem[{Bochkovskiy et~al.(2020)Bochkovskiy, Wang, and Liao}]{bochkovskiy2020yolov4}
Alexey Bochkovskiy, Chien-Yao Wang, and Hong-Yuan~Mark Liao. 2020.
\newblock Yolov4: Optimal speed and accuracy of object detection.
\newblock \emph{arXiv preprint arXiv:2004.10934}.

\bibitem[{Brown et~al.(2020)Brown, Mann, Ryder, Subbiah, Kaplan, Dhariwal, Neelakantan, Shyam, Sastry, Askell et~al.}]{brown2020language}
Tom~B Brown, Benjamin Mann, Nick Ryder, Melanie Subbiah, Jared Kaplan, Prafulla Dhariwal, Arvind Neelakantan, Pranav Shyam, Girish Sastry, Amanda Askell, et~al. 2020.
\newblock Language models are few-shot learners.
\newblock In \emph{Proceedings of the 34th International Conference on Neural Information Processing Systems}, pages 1877--1901.

\bibitem[{Cai et~al.(2019)Cai, Cai, and Wan}]{cai2019multi}
Yitao Cai, Huiyu Cai, and Xiaojun Wan. 2019.
\newblock Multi-modal sarcasm detection in twitter with hierarchical fusion model.
\newblock In \emph{Proceedings of the 57th annual meeting of the association for computational linguistics}, pages 2506--2515.

\bibitem[{Chowdhery et~al.(2022)Chowdhery, Narang, Devlin, Bosma, Mishra, Roberts, Barham, Chung, Sutton, Gehrmann et~al.}]{chowdhery2022palm}
Aakanksha Chowdhery, Sharan Narang, Jacob Devlin, Maarten Bosma, Gaurav Mishra, Adam Roberts, Paul Barham, Hyung~Won Chung, Charles Sutton, Sebastian Gehrmann, et~al. 2022.
\newblock Palm: Scaling language modeling with pathways.
\newblock \emph{arXiv preprint arXiv:2204.02311}.

\bibitem[{Dai et~al.(2023)Dai, Li, Li, Tiong, Zhao, Wang, Li, Fung, and Hoi}]{Dai2023InstructBLIPTG}
Wenliang Dai, Junnan Li, Dongxu Li, Anthony Meng~Huat Tiong, Junqi Zhao, Weisheng Wang, Boyang~Albert Li, Pascale Fung, and Steven C.~H. Hoi. 2023.
\newblock \href {https://api.semanticscholar.org/CorpusID:258615266} {Instructblip: Towards general-purpose vision-language models with instruction tuning}.
\newblock \emph{ArXiv}, abs/2305.06500.

\bibitem[{Davidov et~al.(2010)Davidov, Tsur, and Rappoport}]{davidov2010semi}
Dmitry Davidov, Oren Tsur, and Ari Rappoport. 2010.
\newblock Semi-supervised recognition of sarcasm in twitter and amazon.
\newblock In \emph{Proceedings of the fourteenth conference on computational natural language learning}, pages 107--116.

\bibitem[{Devlin et~al.(2019)Devlin, Chang, Lee, and Toutanova}]{devlin2019bert}
Jacob Devlin, Ming-Wei Chang, Kenton Lee, and Kristina Toutanova. 2019.
\newblock Bert: Pre-training of deep bidirectional transformers for language understanding.
\newblock In \emph{Proceedings of the 2019 Conference of the North American Chapter of the Association for Computational Linguistics: Human Language Technologies, Volume 1 (Long and Short Papers)}, pages 4171--4186.

\bibitem[{Dosovitskiy et~al.(2020)Dosovitskiy, Beyer, Kolesnikov, Weissenborn, Zhai, Unterthiner, Dehghani, Minderer, Heigold, Gelly et~al.}]{dosovitskiy2020image}
Alexey Dosovitskiy, Lucas Beyer, Alexander Kolesnikov, Dirk Weissenborn, Xiaohua Zhai, Thomas Unterthiner, Mostafa Dehghani, Matthias Minderer, Georg Heigold, Sylvain Gelly, et~al. 2020.
\newblock An image is worth 16x16 words: Transformers for image recognition at scale.
\newblock In \emph{International Conference on Learning Representations}.

\bibitem[{Fabbri et~al.(2021)Fabbri, Kry{\'s}ci{\'n}ski, McCann, Xiong, Socher, and Radev}]{fabbri2021summeval}
Alexander~R Fabbri, Wojciech Kry{\'s}ci{\'n}ski, Bryan McCann, Caiming Xiong, Richard Socher, and Dragomir Radev. 2021.
\newblock Summeval: Re-evaluating summarization evaluation.
\newblock \emph{Transactions of the Association for Computational Linguistics}, 9:391--409.

\bibitem[{Girshick(2015)}]{girshick2015fast}
Ross Girshick. 2015.
\newblock Fast r-cnn.
\newblock In \emph{Proceedings of the IEEE international conference on computer vision}, pages 1440--1448.

\bibitem[{Graves and Schmidhuber(2005)}]{graves2005framewise}
Alex Graves and J{\"u}rgen Schmidhuber. 2005.
\newblock Framewise phoneme classification with bidirectional lstm networks.
\newblock In \emph{Proceedings. 2005 IEEE International Joint Conference on Neural Networks, 2005.}, volume~4, pages 2047--2052. IEEE.

\bibitem[{He et~al.(2017)He, Gkioxari, Doll{\'a}r, and Girshick}]{he2017mask}
Kaiming He, Georgia Gkioxari, Piotr Doll{\'a}r, and Ross Girshick. 2017.
\newblock Mask r-cnn.
\newblock In \emph{Proceedings of the IEEE international conference on computer vision}, pages 2961--2969.

\bibitem[{He et~al.(2016)He, Zhang, Ren, and Sun}]{he2016deep}
Kaiming He, Xiangyu Zhang, Shaoqing Ren, and Jian Sun. 2016.
\newblock Deep residual learning for image recognition.
\newblock In \emph{Proceedings of the IEEE conference on computer vision and pattern recognition}, pages 770--778.

\bibitem[{Hu et~al.(2023)Hu, Sheng, Cao, Shi, Li, Wang, and Qi}]{hu2023bad}
Beizhe Hu, Qiang Sheng, Juan Cao, Yuhui Shi, Yang Li, Danding Wang, and Peng Qi. 2023.
\newblock Bad actor, good advisor: Exploring the role of large language models in fake news detection.
\newblock \emph{arXiv preprint arXiv:2309.12247}.

\bibitem[{Huang et~al.(2023)Huang, Kwak, and An}]{huang2023chatgpt}
Fan Huang, Haewoon Kwak, and Jisun An. 2023.
\newblock Is chatgpt better than human annotators? potential and limitations of chatgpt in explaining implicit hate speech.
\newblock In \emph{Companion Proceedings of the ACM Web Conference 2023}, pages 294--297.

\bibitem[{Joshi et~al.(2017)Joshi, Bhattacharyya, and Carman}]{joshi2017automatic}
Aditya Joshi, Pushpak Bhattacharyya, and Mark~J Carman. 2017.
\newblock Automatic sarcasm detection: A survey.
\newblock \emph{ACM Computing Surveys (CSUR)}, 50(5):1--22.

\bibitem[{Joshi et~al.(2018)Joshi, Goel, Bhattacharyya, and Carman}]{joshi2018sarcasm}
Aditya Joshi, Pranav Goel, Pushpak Bhattacharyya, and Mark Carman. 2018.
\newblock Sarcasm target identification: Dataset and an introductory approach.
\newblock In \emph{Proceedings of the Eleventh International Conference on Language Resources and Evaluation (LREC 2018)}.

\bibitem[{Kim(2014)}]{kim2014convolutional}
Yoon Kim. 2014.
\newblock Convolutional neural networks for sentence classification.
\newblock In \emph{Proceedings of the 2014 Conference on Empirical Methods in Natural Language Processing (EMNLP)}. Association for Computational Linguistics.

\bibitem[{Liang et~al.(2021)Liang, Lou, Li, Gui, Yang, and Xu}]{liang2021multi}
Bin Liang, Chenwei Lou, Xiang Li, Lin Gui, Min Yang, and Ruifeng Xu. 2021.
\newblock Multi-modal sarcasm detection with interactive in-modal and cross-modal graphs.
\newblock In \emph{Proceedings of the 29th ACM international conference on multimedia}, pages 4707--4715.

\bibitem[{Liang et~al.(2022)Liang, Lou, Li, Yang, Gui, He, Pei, and Xu}]{liang2022multi}
Bin Liang, Chenwei Lou, Xiang Li, Min Yang, Lin Gui, Yulan He, Wenjie Pei, and Ruifeng Xu. 2022.
\newblock Multi-modal sarcasm detection via cross-modal graph convolutional network.
\newblock In \emph{Proceedings of the 60th Annual Meeting of the Association for Computational Linguistics (Volume 1: Long Papers)}, volume~1, pages 1767--1777. Association for Computational Linguistics.

\bibitem[{Lin et~al.(2024{\natexlab{a}})Lin, Luo, Gao, Ma, Wang, and Yang}]{lin2024explainable}
Hongzhan Lin, Ziyang Luo, Wei Gao, Jing Ma, Bo~Wang, and Ruichao Yang. 2024{\natexlab{a}}.
\newblock Towards explainable harmful meme detection through multimodal debate between large language models.
\newblock In \emph{The ACM Web Conference 2024}, Singapore.

\bibitem[{Lin et~al.(2023{\natexlab{a}})Lin, Luo, Ma, and Chen}]{lin2023beneath}
Hongzhan Lin, Ziyang Luo, Jing Ma, and Long Chen. 2023{\natexlab{a}}.
\newblock Beneath the surface: Unveiling harmful memes with multimodal reasoning distilled from large language models.
\newblock In \emph{The 2023 Conference on Empirical Methods in Natural Language Processing}.

\bibitem[{Lin et~al.(2024{\natexlab{b}})Lin, Luo, Wang, Yang, and Ma}]{lin2024goat}
Hongzhan Lin, Ziyang Luo, Bo~Wang, Ruichao Yang, and Jing Ma. 2024{\natexlab{b}}.
\newblock Goat-bench: Safety insights to large multimodal models through meme-based social abuse.
\newblock \emph{arXiv preprint arXiv:2401.01523}.

\bibitem[{Lin et~al.(2022)Lin, Ma, Chen, Yang, Cheng, and Guang}]{lin2022detect}
Hongzhan Lin, Jing Ma, Liangliang Chen, Zhiwei Yang, Mingfei Cheng, and Chen Guang. 2022.
\newblock Detect rumors in microblog posts for low-resource domains via adversarial contrastive learning.
\newblock In \emph{Findings of the Association for Computational Linguistics: NAACL 2022}, pages 2543--2556.

\bibitem[{Lin et~al.(2021)Lin, Ma, Cheng, Yang, Chen, and Chen}]{lin2021rumor}
Hongzhan Lin, Jing Ma, Mingfei Cheng, Zhiwei Yang, Liangliang Chen, and Guang Chen. 2021.
\newblock Rumor detection on twitter with claim-guided hierarchical graph attention networks.
\newblock In \emph{Proceedings of the 2021 Conference on Empirical Methods in Natural Language Processing}, pages 10035--10047.

\bibitem[{Lin et~al.(2023{\natexlab{b}})Lin, Yi, Ma, Jiang, Luo, Shi, and Liu}]{lin2023zero}
Hongzhan Lin, Pengyao Yi, Jing Ma, Haiyun Jiang, Ziyang Luo, Shuming Shi, and Ruifang Liu. 2023{\natexlab{b}}.
\newblock Zero-shot rumor detection with propagation structure via prompt learning.
\newblock In \emph{Proceedings of the AAAI Conference on Artificial Intelligence}, volume~37, pages 5213--5221.

\bibitem[{Lin et~al.(2014)Lin, Maire, Belongie, Hays, Perona, Ramanan, Doll{\'a}r, and Zitnick}]{lin2014microsoft}
Tsung-Yi Lin, Michael Maire, Serge Belongie, James Hays, Pietro Perona, Deva Ramanan, Piotr Doll{\'a}r, and C~Lawrence Zitnick. 2014.
\newblock Microsoft coco: Common objects in context.
\newblock In \emph{Computer Vision--ECCV 2014: 13th European Conference, Zurich, Switzerland, September 6-12, 2014, Proceedings, Part V 13}, pages 740--755. Springer.

\bibitem[{Liu et~al.(2023{\natexlab{a}})Liu, Li, Wu, and Lee}]{liu2023visual}
Haotian Liu, Chunyuan Li, Qingyang Wu, and Yong~Jae Lee. 2023{\natexlab{a}}.
\newblock Visual instruction tuning.
\newblock \emph{arXiv preprint arXiv:2304.08485}.

\bibitem[{Liu et~al.(2022)Liu, Wang, and Li}]{liu2022towards}
Hui Liu, Wenya Wang, and Haoliang Li. 2022.
\newblock Towards multi-modal sarcasm detection via hierarchical congruity modeling with knowledge enhancement.
\newblock In \emph{Proceedings of the 2022 Conference on Empirical Methods in Natural Language Processing}, pages 4995--5006.

\bibitem[{Liu et~al.(2023{\natexlab{b}})Liu, Zeng, Ren, Li, Zhang, Yang, Li, Yang, Su, Zhu et~al.}]{liu2023grounding}
Shilong Liu, Zhaoyang Zeng, Tianhe Ren, Feng Li, Hao Zhang, Jie Yang, Chunyuan Li, Jianwei Yang, Hang Su, Jun Zhu, et~al. 2023{\natexlab{b}}.
\newblock Grounding dino: Marrying dino with grounded pre-training for open-set object detection.
\newblock \emph{arXiv preprint arXiv:2303.05499}.

\bibitem[{Liu et~al.(2019)Liu, Ott, Goyal, Du, Joshi, Chen, Levy, Lewis, Zettlemoyer, and Stoyanov}]{liu2019roberta}
Yinhan Liu, Myle Ott, Naman Goyal, Jingfei Du, Mandar Joshi, Danqi Chen, Omer Levy, Mike Lewis, Luke Zettlemoyer, and Veselin Stoyanov. 2019.
\newblock Roberta: A robustly optimized bert pretraining approach.
\newblock \emph{arXiv preprint arXiv:1907.11692}.

\bibitem[{Liu et~al.(2021)Liu, Lin, Cao, Hu, Wei, Zhang, Lin, and Guo}]{liu2021swin}
Ze~Liu, Yutong Lin, Yue Cao, Han Hu, Yixuan Wei, Zheng Zhang, Stephen Lin, and Baining Guo. 2021.
\newblock Swin transformer: Hierarchical vision transformer using shifted windows.
\newblock In \emph{Proceedings of the IEEE/CVF international conference on computer vision}, pages 10012--10022.

\bibitem[{Luo et~al.(2023)Luo, Xu, Zhao, Sun, Geng, Hu, Tao, Ma, Lin, and Jiang}]{wizardcoder}
Ziyang Luo, Can Xu, Pu~Zhao, Qingfeng Sun, Xiubo Geng, Wenxiang Hu, Chongyang Tao, Jing Ma, Qingwei Lin, and Daxin Jiang. 2023.
\newblock \href {https://doi.org/10.48550/ARXIV.2306.08568} {Wizardcoder: Empowering code large language models with evol-instruct}.
\newblock \emph{CoRR}, abs/2306.08568.

\bibitem[{Mehri and Eskenazi(2020)}]{mehri2020unsupervised}
Shikib Mehri and Maxine Eskenazi. 2020.
\newblock Unsupervised evaluation of interactive dialog with dialogpt.
\newblock In \emph{Proceedings of the 21th Annual Meeting of the Special Interest Group on Discourse and Dialogue}, pages 225--235.

\bibitem[{Minderer et~al.(2022)Minderer, Gritsenko, Stone, Neumann, Weissenborn, Dosovitskiy, Mahendran, Arnab, Dehghani, Shen et~al.}]{minderer2022simple}
Matthias Minderer, Alexey Gritsenko, Austin Stone, Maxim Neumann, Dirk Weissenborn, Alexey Dosovitskiy, Aravindh Mahendran, Anurag Arnab, Mostafa Dehghani, Zhuoran Shen, et~al. 2022.
\newblock Simple open-vocabulary object detection.
\newblock In \emph{European Conference on Computer Vision}, pages 728--755. Springer.

\bibitem[{Molla and Joshi(2019)}]{molla2019overview}
Diego Molla and Aditya Joshi. 2019.
\newblock Overview of the 2019 alta shared task: Sarcasm target identification.
\newblock In \emph{Proceedings of the the 17th annual workshop of the Australasian language technology association}, pages 192--196.

\bibitem[{OpenAI(2023)}]{OpenAI2023GPT4TR}
OpenAI. 2023.
\newblock \href {https://api.semanticscholar.org/CorpusID:257532815} {Gpt-4 technical report}.
\newblock \emph{ArXiv}, abs/2303.08774.

\bibitem[{Pan et~al.(2020)Pan, Lin, Fu, Qi, and Wang}]{pan2020modeling}
Hongliang Pan, Zheng Lin, Peng Fu, Yatao Qi, and Weiping Wang. 2020.
\newblock Modeling intra and inter-modality incongruity for multi-modal sarcasm detection.
\newblock In \emph{Findings of the Association for Computational Linguistics: EMNLP 2020}, pages 1383--1392.

\bibitem[{Parameswaran et~al.(2019)Parameswaran, Trotman, Liesaputra, and Eyers}]{parameswaran2019detecting}
Pradeesh Parameswaran, Andrew Trotman, Veronica Liesaputra, and David Eyers. 2019.
\newblock Detecting target of sarcasm using ensemble methods.
\newblock In \emph{Proceedings of the the 17th annual workshop of the Australasian language technology association}, pages 197--203.

\bibitem[{Patro et~al.(2019)Patro, Bansal, and Mukherjee}]{patro2019deep}
Jasabanta Patro, Srijan Bansal, and Animesh Mukherjee. 2019.
\newblock A deep-learning framework to detect sarcasm targets.
\newblock In \emph{Proceedings of the 2019 conference on empirical methods in natural language processing and the 9th international joint conference on natural language processing (emnlp-ijcnlp)}, pages 6336--6342.

\bibitem[{Qin et~al.(2023)Qin, Huang, Chen, Cai, Zhang, Liang, Che, and Xu}]{qin2023mmsd2}
Libo Qin, Shijue Huang, Qiguang Chen, Chenran Cai, Yudi Zhang, Bin Liang, Wanxiang Che, and Ruifeng Xu. 2023.
\newblock Mmsd2. 0: Towards a reliable multi-modal sarcasm detection system.
\newblock In \emph{Findings of the Association for Computational Linguistics: ACL 2023}, pages 10834--10845.

\bibitem[{Radford et~al.(2021)Radford, Kim, Hallacy, Ramesh, Goh, Agarwal, Sastry, Askell, Mishkin, Clark et~al.}]{radford2021learning}
Alec Radford, Jong~Wook Kim, Chris Hallacy, Aditya Ramesh, Gabriel Goh, Sandhini Agarwal, Girish Sastry, Amanda Askell, Pamela Mishkin, Jack Clark, et~al. 2021.
\newblock Learning transferable visual models from natural language supervision.
\newblock In \emph{International conference on machine learning}, pages 8748--8763. PMLR.

\bibitem[{Raffel et~al.(2020)Raffel, Shazeer, Roberts, Lee, Narang, Matena, Zhou, Li, and Liu}]{raffel2020exploring}
Colin Raffel, Noam Shazeer, Adam Roberts, Katherine Lee, Sharan Narang, Michael Matena, Yanqi Zhou, Wei Li, and Peter~J Liu. 2020.
\newblock Exploring the limits of transfer learning with a unified text-to-text transformer.
\newblock \emph{The Journal of Machine Learning Research}, 21(1):5485--5551.

\bibitem[{Rezatofighi et~al.(2019)Rezatofighi, Tsoi, Gwak, Sadeghian, Reid, and Savarese}]{rezatofighi2019generalized}
Hamid Rezatofighi, Nathan Tsoi, JunYoung Gwak, Amir Sadeghian, Ian Reid, and Silvio Savarese. 2019.
\newblock Generalized intersection over union: A metric and a loss for bounding box regression.
\newblock In \emph{Proceedings of the IEEE/CVF conference on computer vision and pattern recognition}, pages 658--666.

\bibitem[{Riloff et~al.(2013)Riloff, Qadir, Surve, De~Silva, Gilbert, and Huang}]{riloff2013sarcasm}
Ellen Riloff, Ashequl Qadir, Prafulla Surve, Lalindra De~Silva, Nathan Gilbert, and Ruihong Huang. 2013.
\newblock Sarcasm as contrast between a positive sentiment and negative situation.
\newblock In \emph{Proceedings of the 2013 conference on empirical methods in natural language processing}, pages 704--714.

\bibitem[{Schifanella et~al.(2016)Schifanella, De~Juan, Tetreault, and Cao}]{schifanella2016detecting}
Rossano Schifanella, Paloma De~Juan, Joel Tetreault, and Liangliang Cao. 2016.
\newblock Detecting sarcasm in multimodal social platforms.
\newblock In \emph{Proceedings of the 24th ACM international conference on Multimedia}, pages 1136--1145.

\bibitem[{Team et~al.(2023)Team, Anil, Borgeaud, Wu, Alayrac, Yu, Soricut, Schalkwyk, Dai, Hauth et~al.}]{team2023gemini}
Gemini Team, Rohan Anil, Sebastian Borgeaud, Yonghui Wu, Jean-Baptiste Alayrac, Jiahui Yu, Radu Soricut, Johan Schalkwyk, Andrew~M Dai, Anja Hauth, et~al. 2023.
\newblock Gemini: A family of highly capable multimodal models.
\newblock \emph{arXiv preprint arXiv:2312.11805}.

\bibitem[{Terven and Cordova-Esparza(2023)}]{terven2023comprehensive}
Juan Terven and Diana Cordova-Esparza. 2023.
\newblock A comprehensive review of yolo: From yolov1 to yolov8 and beyond.
\newblock \emph{arXiv preprint arXiv:2304.00501}.

\bibitem[{Tian et~al.(2023)Tian, Xu, Zhang, and Mao}]{tian2023dynamic}
Yuan Tian, Nan Xu, Ruike Zhang, and Wenji Mao. 2023.
\newblock Dynamic routing transformer network for multimodal sarcasm detection.
\newblock In \emph{Proceedings of the 61st Annual Meeting of the Association for Computational Linguistics (Volume 1: Long Papers)}, pages 2468--2480.

\bibitem[{Touvron et~al.(2023{\natexlab{a}})Touvron, Lavril, Izacard, Martinet, Lachaux, Lacroix, Rozi{\`e}re, Goyal, Hambro, Azhar et~al.}]{touvron2023llama}
Hugo Touvron, Thibaut Lavril, Gautier Izacard, Xavier Martinet, Marie-Anne Lachaux, Timoth{\'e}e Lacroix, Baptiste Rozi{\`e}re, Naman Goyal, Eric Hambro, Faisal Azhar, et~al. 2023{\natexlab{a}}.
\newblock Llama: Open and efficient foundation language models.
\newblock \emph{arXiv preprint arXiv:2302.13971}.

\bibitem[{Touvron et~al.(2023{\natexlab{b}})Touvron, Martin, Stone, Albert, Almahairi, Babaei, Bashlykov, Batra, Bhargava, Bhosale et~al.}]{touvron2023llama2}
Hugo Touvron, Louis Martin, Kevin Stone, Peter Albert, Amjad Almahairi, Yasmine Babaei, Nikolay Bashlykov, Soumya Batra, Prajjwal Bhargava, Shruti Bhosale, et~al. 2023{\natexlab{b}}.
\newblock Llama 2: Open foundation and fine-tuned chat models.
\newblock \emph{arXiv preprint arXiv:2307.09288}.

\bibitem[{Wang et~al.(2022)Wang, Sun, Liu, Shao, and Zheng}]{wang2022multimodal}
Jiquan Wang, Lin Sun, Yi~Liu, Meizhi Shao, and Zengwei Zheng. 2022.
\newblock Multimodal sarcasm target identification in tweets.
\newblock In \emph{Proceedings of the 60th Annual Meeting of the Association for Computational Linguistics (Volume 1: Long Papers)}, pages 8164--8175.

\bibitem[{Wang et~al.(2023)Wang, Lv, Yu, Hong, Qi, Wang, Ji, Yang, Zhao, Song et~al.}]{wang2023cogvlm}
Weihan Wang, Qingsong Lv, Wenmeng Yu, Wenyi Hong, Ji~Qi, Yan Wang, Junhui Ji, Zhuoyi Yang, Lei Zhao, Xixuan Song, et~al. 2023.
\newblock Cogvlm: Visual expert for pretrained language models.
\newblock \emph{arXiv preprint arXiv:2311.03079}.

\bibitem[{Wei et~al.(2022)Wei, Wang, Schuurmans, Bosma, Xia, Chi, Le, Zhou et~al.}]{wei2022chain}
Jason Wei, Xuezhi Wang, Dale Schuurmans, Maarten Bosma, Fei Xia, Ed~H Chi, Quoc~V Le, Denny Zhou, et~al. 2022.
\newblock Chain-of-thought prompting elicits reasoning in large language models.
\newblock In \emph{Advances in Neural Information Processing Systems}.

\bibitem[{Xiong et~al.(2019)Xiong, Zhang, Zhu, and Yang}]{xiong2019sarcasm}
Tao Xiong, Peiran Zhang, Hongbo Zhu, and Yihui Yang. 2019.
\newblock Sarcasm detection with self-matching networks and low-rank bilinear pooling.
\newblock In \emph{The world wide web conference}, pages 2115--2124.

\bibitem[{Xu et~al.(2023)Xu, Sun, Zheng, Geng, Zhao, Feng, Tao, and Jiang}]{wizardlm}
Can Xu, Qingfeng Sun, Kai Zheng, Xiubo Geng, Pu~Zhao, Jiazhan Feng, Chongyang Tao, and Daxin Jiang. 2023.
\newblock \href {https://doi.org/10.48550/ARXIV.2304.12244} {Wizardlm: Empowering large language models to follow complex instructions}.
\newblock \emph{CoRR}, abs/2304.12244.

\bibitem[{Xu et~al.(2020)Xu, Zeng, and Mao}]{xu2020reasoning}
Nan Xu, Zhixiong Zeng, and Wenji Mao. 2020.
\newblock Reasoning with multimodal sarcastic tweets via modeling cross-modality contrast and semantic association.
\newblock In \emph{Proceedings of the 58th annual meeting of the association for computational linguistics}, pages 3777--3786.

\bibitem[{Yang et~al.(2023{\natexlab{a}})Yang, Gao, Ma, Lin, and Yang}]{yang2023wsdms}
Ruichao Yang, Wei Gao, Jing Ma, Hongzhan Lin, and Zhiwei Yang. 2023{\natexlab{a}}.
\newblock Wsdms: Debunk fake news via weakly supervised detection of misinforming sentences with contextualized social wisdom.
\newblock In \emph{Proceedings of the 2023 Conference on Empirical Methods in Natural Language Processing}, pages 1525--1538.

\bibitem[{Yang et~al.(2023{\natexlab{b}})Yang, Li, Lin, Wang, Lin, Liu, and Wang}]{yang2023dawn}
Zhengyuan Yang, Linjie Li, Kevin Lin, Jianfeng Wang, Chung-Ching Lin, Zicheng Liu, and Lijuan Wang. 2023{\natexlab{b}}.
\newblock The dawn of lmms: Preliminary explorations with gpt-4v (ision).
\newblock \emph{arXiv preprint arXiv:2309.17421}, 9(1).

\bibitem[{Yin et~al.(2009)Yin, Xue, Hong, Davison, Kontostathis, Edwards et~al.}]{yin2009detection}
Dawei Yin, Zhenzhen Xue, Liangjie Hong, Brian~D Davison, April Kontostathis, Lynne Edwards, et~al. 2009.
\newblock Detection of harassment on web 2.0.
\newblock \emph{Proceedings of the Content Analysis in the WEB}, 2(0):1--7.

\bibitem[{Zeng et~al.(2022)Zeng, Liu, Du, Wang, Lai, Ding, Yang, Xu, Zheng, Xia et~al.}]{zeng2022glm}
Aohan Zeng, Xiao Liu, Zhengxiao Du, Zihan Wang, Hanyu Lai, Ming Ding, Zhuoyi Yang, Yifan Xu, Wendi Zheng, Xiao Xia, et~al. 2022.
\newblock Glm-130b: An open bilingual pre-trained model.
\newblock In \emph{The Eleventh International Conference on Learning Representations}.

\bibitem[{Zhang et~al.(2022)Zhang, Li, Liu, Zhang, Su, Zhu, Ni, and Shum}]{zhang2022dino}
Hao Zhang, Feng Li, Shilong Liu, Lei Zhang, Hang Su, Jun Zhu, Lionel~M Ni, and Heung-Yeung Shum. 2022.
\newblock Dino: Detr with improved denoising anchor boxes for end-to-end object detection.
\newblock \emph{arXiv preprint arXiv:2203.03605}.

\end{thebibliography}
\newpage
\appendix



\section{Datasets}
\begin{table}[]\centering
\resizebox{0.7\linewidth}{!}{
\begin{tabular}{lccc}
\toprule
MMSD2.0   & Train & Dev  & Test \\ \hline
Sentences & 19816 & 2410 & 2409 \\
Positive  & 9572  & 1042 & 1037 \\
Negative  & 10240 & 1368 & 1372 \\ \toprule
\end{tabular}}
\caption{Statistics of the MMSD2.0 dataset.}
\label{statistic_mmsd2.0}
\end{table}

\begin{table}[]\centering
\resizebox{\linewidth}{!}{
\begin{tabular}{lccc}
\toprule
MSTI/MSTI2.0 & Train     & Dev     & Test    \\
\hline
Textual ST   & 2266/2830 & 360/502 & 367/517 \\
Visual ST    & 1614/860  & 402/239 & 419/220 \\
Total        & 3546      & 727     & 742   \\ \toprule
\end{tabular}}
\caption{Statistic comparison between MSTI and MSTI2.0.}
\label{statistic_msti2.0}
\end{table}

\begin{table*}[]
\resizebox{\textwidth}{!}{
\begin{tabular}{lcccccccccc}
\toprule
\multicolumn{1}{c}{}     & \multicolumn{5}{c}{Dev}               & \multicolumn{5}{c}{Test}              \\ \cmidrule(lr){2-6} \cmidrule(l){7-11} 
                         & EM    & F1    & AP    & AP50  & AP75  & EM    & F1    & AP    & AP50  & AP75  \\ \hline
MSTI-CB (reported)   & 34.20 & 44.90 & 32.10 & 52.30 & 34.20 & 35.00 & 45.80 & 32.30 & 51.80 & 34.00 \\
MSTI-CB (reproduced) & 35.51 & 43.09 & 00.74 & 02.40 & 00.12 & 33.15 & 41.22 & 01.20 & 04.50 & 00.03 \\
CofiPara-\textsc{Msti} & 39.37 & 45.90 & 04.68 & 09.79 & 03.95 & 40.04 & 44.43 & 04.22 & 09.62 & 03.54 \\ 
\toprule
\end{tabular}}
\caption{The gap between the reported MSTI-CB results in the original paper~\citep{wang2022multimodal} and the reproduced MSTI-CB results by the released official codes on the original MSTI dataset. And we further provide the results of our proposed approach on the original MSTI dataset.}
\label{ori_msti}
\end{table*}

\begin{figure}
    \centering
    \includegraphics[width=\linewidth, scale=1.00]{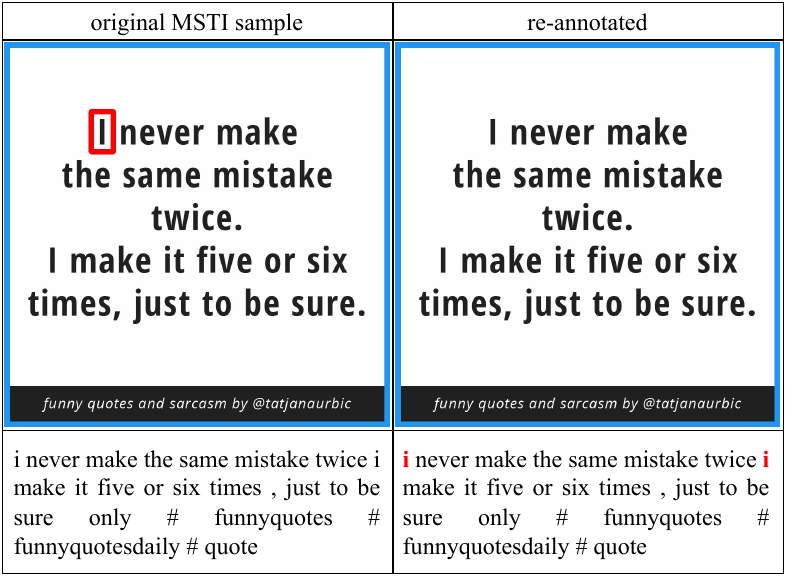}
    \caption{An example of re-annotated samples.} 
    \label{ocr_example}
\end{figure}

The statistics of MMSD2.0 and MSTI2.0 are shown in Table~\ref{statistic_mmsd2.0} and Table~\ref{statistic_msti2.0}, respectively. In our experiments, we noted a significant discrepancy in the visual performance of the MSTI\footnote{https://github.com/wjq-learning/MSTI} baseline, trained using identical settings from their repository, compared to results reported in their original publication~\citep{wang2022multimodal}, as illustrated in Table~\ref{ori_msti}. We dived into the quality of the MSTI dataset is poor due to the original MSTI dataset's substantial inclusion of visual labels targeting optical characters, as shown in the left column in Figure \ref{ocr_example}. We contend that labeling such entities as visual targets is not fitting for several reasons: 1) First, identifying optical character targets in synthetic images, such as memes or screenshots, fundamentally leans towards a Natural Language Processing (NLP) challenge. Applying visual metrics like AP50 to assess a task that is inherently textual in nature is clearly impractical. 2) Second, the recurrence of certain optical characters within a single image, such as the words ``I'', ``you'', or ``Trump'', often goes partially annotated in the original MSTI dataset, where typically only the first occurrence is marked, while subsequent instances are overlooked. This approach to annotation is also questionable. The statistic of optical character targets in the original MSTI dataset is listed in Table \ref{sta_ocr}, where \textit{OCR targets} denotes the optical character targets that consist of 48.53\% of the visual labels.

We re-annotated the original dataset in the following steps: 1) First, by calculating the area ratios of truth-bounding boxes to images, we first roughly filter out samples that might contain OCR targets with a threshold of 0.15. 2) Then, by using the EasyOCR\footnote{\url{https://github.com/JaidedAI/EasyOCR}} tool, we extract texts in these images and their corresponding bounding boxes. 3) To further adjust the extracted results, we use LLaVA for result correction. Specifically, we feed the extracted texts and images into LLaVA, and prompt the model to check if the texts match the characters in the images. 4) Moreover, we applied a manual validation to double-check the corrected results, and finally concatenate the results to the original MSTI texts. The original bounding box labels are then deleted after being transformed into textual forms.

An example of the original and re-annotated sample is shown in Figure~\ref{ocr_example}. The bounding box is labeled in the red rectangle, and textual labels are highlighted in red bold. The original sample on the left neglected the second “I” in the image, and the word “I” in the text also remains unlabeled, whereas in the re-annotated version, we removed the bounding box and labeled the target word in textual form.


\begin{table}[]\centering
\resizebox{0.7\linewidth}{!}{
\begin{tabular}{lccc}
\toprule
                & Train & Dev & Test \\ \hline
OCR targets     & 754   & 163 & 199  \\
non-OCR targets & 860   & 239 & 220  \\ \toprule
\end{tabular}
}
\caption{Statistics of optical labels in the original MSTI dataset.}
\label{sta_ocr}
\end{table}

\section{Baselines}
We compare our model with the following baselines for multimodal sarcasm detection: 1) \textsf{TextCNN}~\citep{kim2014convolutional}:  a neural network based on CNN for textual classification; 2) \textsf{Bi-LSTM}~\citep{graves2005framewise}: a bi-directional long short-term memory network for textual classification; 3) \textsf{SMSD}~\citep{xiong2019sarcasm}: a self-matching network to capture the incongruity information in sentences by exploring word-to-word interactions; 4) \textsf{RoBERTa}~\citep{liu2019roberta}: an optimized version of the BERT~\citep{devlin2019bert} language model; 5) \textsf{ResNet}~\citep{he2016deep}: we employ the image embedding of the pooling layer of ResNet for sarcasm classification; 6) \textsf{ViT}~\citep{dosovitskiy2020image}: a neural network model that applies the transformer architecture to computer vision tasks; 7) \textsf{HFM}~\citep{cai2019multi}: a hierarchical fusion model for multimodal sarcasm detection; 8) \textsf{Att-BERT}~\citep{pan2020modeling}: two attention mechanisms is devised to model the textonly and cross-modal incongruity, respectively; 9) \textsf{CMGCN}~\citep{liang2022multi}: a fine-grained cross-modal graph architecture based on attribute-object pairs of image objects to capture sarcastic clues; 10) \textsf{HKE}~\citep{liu2022towards}: a hierarchical graph-based framework to model atomic-level and compositionlevel congruity; 11) \textsf{DynRT-Net}~\citep{tian2023dynamic}: a dynamic routing transformer network via adapting dynamic paths to hierarchical co-attention between image and text modalities; 12) \textsf{Multi-view CLIP}~\citep{qin2023mmsd2}: a multimodal model based on CLIP~\citep{radford2021learning} to capture sarcasm cues from different perspectives, including image view, text view, and image-text interactions view. 

We adopt Accuracy, F1 score, Precision, and Recall to evaluate the MSD performance. Although we make a comparison with the MSD baselines for the illustration of our universal framework, please note that the MSD task is just an auxiliary task but not our goal task.

To evaluate our model in multimodal sarcasm target identification, we compare the following state-of-the-art MSTI systems: 1) \textsf{BERT-Base}~\citep{devlin2019bert}: the base version of a classical auto-encoding pre-trained language model is applied to the sequence tagging task; 2) \textsf{BERT-Large}: the large version of the BERT model; 3) \textsf{Mask R-CNN}~\citep{he2017mask}: an extension of Faster R-CNN~\citep{girshick2015fast} that adds a branch for predicting segmentation masks on each Region of Interest, enabling pixel-level object instance segmentation; 4) \textsf{YOLOv8}~\citep{terven2023comprehensive}: an advanced version of the YOLO (You Only Look Once) series, designed for real-time object detection with improved accuracy and speed; 5) \textsf{OWL-ViT}~\citep{minderer2022simple}: a pre-trained vision-language model designed for open-vocabulary object detection; 6) \textsf{Grounding DINO}~\citep{liu2023grounding}: a pre-trained vision-language model that could act as a powerful zero-shot object detector in the open set; 7) \textsf{MSTI-RB}~\citep{wang2022multimodal}: a variant of the MSTI model with ResNet152 as the backbone and BERT-Base as the language model; 8) \textsf{MSTI-VB}: a variant of the MSTI model with VGG19 as the backbone and BERT-Base as the language model; 9) \textsf{MSTI-CB}: a variant of the MSTI model with CSPDarkNet53 as the backbone and BERT-Base as the language model; 10) \textsf{MSTI-CL}: a variant of the MSTI model with CSPDarkNet53 as the backbone and BERT-Large as the language model.

We use Exact Match (EM) accuracy~\citep{joshi2018sarcasm} and F1 score~\citep{molla2019overview} as evaluation metrics of textual sarcasm target identification; and Average Precision (AP)~\citep{lin2014microsoft}, i.e., the COCO-style AP, AP50, and AP75, as the metrics for visual sarcasm target identification.

Please note that due to our inability to evaluate all potential models, we can only select the representative state-of-the-art baselines for the text-modality, image-modality, and multimodal types.

\section{Implementation Details}
In our experiments, we utilize two representative LMMs, LLaVA ~\citep{liu2023visual} and Qwen-VL ~\citep{bai2023qwen} to generate rationales, for their excellence in tasks like multimodal question answering and reasoning. Specifically, we implement the ``LLaVA-1.5 13B'' and ``v1.0.0'' versions for LLaVA and Qwen-VL, respectively. Note that the choice of LMMs is orthogonal to our proposed paradigm, which can be easily replaced by any existing LMM without further modification.
Our CofiPara utilizes T5 architecture ~\citep{raffel2020exploring} as our text encoder and decoder backbone. Specifically, we use the “flan-t5-base” version to initialize our language model. For image encoder, bi-attention block, and cross-modality image decoder structure, we utilize the same setting as GroundingDINO-T \footnote{\url{ https://github.com/IDEA-Research/GroundingDINO}}, and initialize using SwinT\_OGC checkpoint in the same repository, where we freeze image encoder parameters during the whole training process. For text decoding, we adapt a text-to-image cross-attention module that matches text query with image key and value, which is randomly initialized.
Hyper-parameters used during training are listed in Table \ref{hyperp}, where these hyper-parameters are shared during both pre-training and fine-tuning process. Our results are averaged over 10 random runs. All of our experiments are conducted on a single NVIDIA RTX A6000 GPU. The size of our model is about 320M, among which 295M parameters are trainable. Training takes about 1 hour per epoch at the pre-training stage, and about 30 minutes per epoch at the fine-tuning stage.
For MSTI baseline implementation, we use the exact same settings for training as reported in their paper ~\citep{wang2022multimodal}. As for the OWL-ViT baseline, we use the specific version of “OWLv1 CLIP ViT-B/32”, and train the model for 10 epochs using a batch size of 8, a learning rate of 5e-5. For the Grounding DINO baseline, we also implement the SwinT\_OGC version, and we report the zero-shot results as the released version only supports zero-shot inference in their current repository without training codes.

\begin{table}[]\centering
\resizebox{0.8\linewidth}{!}{
\begin{tabular}{lc}
\toprule
Hyper-Parameters & \\ \hline\hline
Epoch         & 10   \\
Batch Size    & 4    \\
Learning Rate & 5e-5 \\
Optimizer     & Adam \\
Adam eps      & 1e-8 \\
Image Size    & 600  \\
L1 coefficient $\alpha$      & 0.2\\
GIOU coefficient $\beta$       & 1e-3\\ 
Classification coefficient $\gamma$      & 0.1\\   
Number of text decoder layers $L$ & 12 \\
Number of image decoder layers $K$ &  6 \\
\toprule
\end{tabular}}
\caption{Hyper-parameters used during pre-training and fine-tuning.}
\label{hyperp}
\end{table}

\section{CofiPara Algorithm}
Algorithm \ref{algoritm1} presents the coarse-to-fine training paradigm of our approach.

First, we perform divergent thinking strategy on two datasets $D_1$ and $D_2$ for the pre-training and fine-tuning stages, respectively. For each MSD sample $ M = \{I,T\} $ in $D_1$, we generate a rationale pair including a negative (i.e., \texttt{non-sarcastic}) and a positive (i.e., \texttt{sarcastic}) one with potential label $*$ that explicates the input as sarcastic/non-sarcastic. For MSTI samples in $D_2$, we only generate a single rationale $r_{pos}$ that explains why the input sample is sarcastic. Then, we utilize the dataset $D_1$ for coarser-grained pre-training, where the model is trained to identify if the input sample $M$ is sarcastic or not, and use the cross-entropy loss as the training loss. After $epoches$ of iteration for pre-training, we further fine-tune our model on the dataset $D_2$, to identify textual and visual sarcasm targets in the inputs. In the fine-tuning stage, the model outputs the predicted textual and visual target $\hat{y}_{text},\hat{y}_{img}$, with which we calculate text and image loss as shown in Equation \ref{eq6} and Equation \ref{eq_imgloss}. Finally, by adding up text loss $ \hat{\mathcal{L}}_{text} $ and image loss $ \mathcal{L}_{img} $, we optimize our model with the loss $\mathcal{L}$.

\begin{algorithm}
 \SetAlgoNoLine 
 \caption{Model Training}
  \KwIn{MSD dataset $D_1$, MSTI dataset $D_2$, template $p$, $LMM$, our model $f$}
    \textbf{Rationale Generation:} \\
    \For{$M = \{I,T\} \in D_1,D_2$ }
    {   
        \If{$M \in D_1$}
            {$r_{pos}, r_{neg} = LMM(p(I,T,$*$))$\\
            $\hat{T} = \{T, r_{pos}, r_{neg}\}$}
        \Else
            {$r_{pos} = LMM(p(I,T,$*=$\texttt{sarcastic}))$ \\
            $\hat{T} = \{T, r_{pos}\}$ }  
    }
    \textbf{Pre-training Stage:} \\
    \While{$epoch < epoches$}
    {
        \For{$M=\{I,\hat{T},y\} \in D_1$ }
        {
        $\hat{y} = f(I,\hat{T})$ \\
        $\mathcal{L}_{text} = CE(\hat{y}, y)$
        }
    }
    \textbf{Fine-tuning Stage:} \\
    \While{$epoch < epoches$}
    {
        \For{$M=\{I,\hat{T},y\} \in D_2$ }
        {
        $\hat{y}_{text},\hat{y}_{img} = f(I,\hat{T})$ \\
        calculate $\hat{\mathcal{L}}_{text}$ as Equation \ref{eq6}.\\
        calculate $\mathcal{L}_{img}$ as Equation \ref{eq_imgloss}.\\
        $\mathcal{L} = \mathcal{L}_{img} + \hat{\mathcal{L}}_{text}$
        }
    }
    \label{algoritm1}
\end{algorithm}

\section{More Results of Sarcasm Detection Performance}
\begin{table}[] \small
\centering
\resizebox{0.9\linewidth}{!}{\begin{tabular}{lcccc}
\toprule
\multicolumn{1}{c}{}       & \multicolumn{1}{l}{Acc.} & P     & R     & F1    \\ \hline
TextCNN                  & 71.61                    & 64.62 & 75.22 & 69.52 \\
Bi-LSTM                    & 72.48                    & 68.02 & 68.08 & 68.05 \\
SMSD                       & 73.56                    & 68.45 & 71.55 & 69.97 \\
RoBERTa                    & 79.66                    & 76.74 & 75.70 & 76.21 \\ \hline
ResNet                     & 65.50                    & 61.17 & 54.39 & 57.58 \\
ViT                       & 72.02                    & 65.26 & 74.83 & 69.72 \\
HFM                        & 70.57                    & 64.84 & 69.50 & 66.88 \\ \hline
Att-BERT                   & 80.03                    & 76.28 & 77.82 & 77.04 \\
CMGCN                      & 79.83                    & 75.82 & 78.01 & 76.90 \\
HKE                      & 76.50                    & 73.48 & 71.07 & 72.25 \\
DynRT                      & 72.06                    & 71.79 & 72.18 & 71.98 \\
Multi-view CLIP           & 84.31                    & 79.66 & 85.34 & 82.40 \\ \hline
CofiPara-\textsc{Msd}               & \textbf{85.70}                    & \textbf{85.96} & \textbf{85.55} & \textbf{85.89} \\
\toprule 
\end{tabular}}
\caption{Multimodal sarcasm detection results with text-modality, image-modality, and multimodal baselines.}
\label{msd_res_full}
\end{table}

Table \ref{msd_res_full} illustrates the performance (\%) of our proposed method versus all the compared sarcasm detection baselines on the auxiliary task MSD ($p < 0.05$ under t-test). From the results, we have the following observations: 1) The multimodal methods in the third group generally outperform the unimodal methods of text and image modality in the first and second groups, respectively. 2) Compared to graph-based methods such as CMGCN and HKE and routing-based DynRT, models like RoBERTa and Att-BERT that rely on semantic understanding have much better performance, indicating that this task requires models to capture deep semantic information rather than superficial attributes. 3) Multi-view CLIP shows an overall advantage in its ability to align textual and visual features, and the isomorphic structures of text and image encoder also contribute to its superiority. 4) Our proposed CofiPara-\textsc{Msd}, as a by-product of the coarse-to-fine framework, surpasses the leading baseline by 1.39\% and 3.49\% on the accuracy and F1 score, additionally demonstrating a more balanced performance in terms of recall and precision, despite not primarily targeting the MSD task. The distinctive advantage of our model lies in the fact that while all the baselines solely focus on recognition, our model is equipped with rationales from divergent thinking with LMMs, which empowers our model to effectively uncover sarcastic content by adeptly leveraging the interplay between seemingly unrelated textual and visual elements within sarcasm.

\section{LMM for MSD}
\begin{table}[]
\resizebox{\linewidth}{!}{\begin{tabular}{lcccc}
\toprule
\multicolumn{1}{c}{}         & Acc.  & P     & R     & F1    \\ \hline
LLaVA (Zero-shot)            & 51.06 & 40.09 & 46.40 & 43.02 \\
LLaVA (CoT)                  & 48.69 & 40.93 & 65.17 & 50.28 \\
LLaVA (Divergent-thinking)   & 80.41 & 82.22 & 82.55 & 80.40 \\
LLaVA (CofiPara)             & 84.51 & 84.58 & 84.17 & 84.43 \\ 
\hline
Qwen-VL (Zero-shot)          & 76.63 & 67.44 & 68.53 & 69.03 \\
Qwen-VL (CoT)                & 68.86 & 66.82 & 68.67 & 66.96 \\
Qwen-VL (Divergent-thinking) & 75.88 & 73.68 & 77.19 & 74.12 \\
Qwen-VL (CofiPara)           & 85.70 & 85.96 & 85.55 & 85.89 \\ \hline
CofiPara-\textsc{Msd} w/o LMM & 81.61                    & 81.92 & 81.36 & 81.46 \\
\toprule
\end{tabular}}
\caption{Effect of different LMMs on MMSD2.0 dataset.}
\label{lmm_res}
\end{table}

Table~\ref{lmm_res} shows the effect of different LMMs (i.e., LLaVA~\citep{liu2023visual} and Qwen-VL~\citep{bai2023qwen}) on the auxiliary task MSD. The `Divergent-Thinking' prompting strategy could effectively enhance the multimodal sarcasm detection performance of LMMs~\citep{lin2024goat}, especially LLaVA, which suggests that the conflicting rationales generation from the divergent thinking is a reasonable way to optimize the reasoning chains for LMMs applied to the MSD task. 
We notice that our model performance dropped on accuracy and F1 score with LMM-generated rationales removed. This suggests that LMM-generated rationales do provide supportive knowledge for detecting sarcasm, and that our model can effectively filter out useful information from contrary rationales. This design makes our model robust in the following fine-tuning stage for MSTI, which could alleviate the negative impact posed by potential noise in the rationale from only the sarcastic perspective.

\section{LMM for MSTI}

\begin{table*}[]
\resizebox{\textwidth}{!}{
\begin{tabular}{lcccccccccc}
\toprule
\multicolumn{1}{c}{}              & \multicolumn{5}{c}{Dev}               & \multicolumn{5}{c}{Test}              \\ \cmidrule(lr){2-6} \cmidrule(l){7-11}
                                  & EM    & F1    & AP    & AP50  & AP75  & EM    & F1    & AP    & AP50  & AP75  \\ \hline
LLaVA (Zero-shot)                            & 16.05 & 00.14 & /     & /     & /     & 13.20 & 00.40 & /     & /     & /     \\
LLaVA (CoT)  & 16.23 & 00.01 & /     & /     & /     & 13.20 & 00.53 & /     & /     & /     \\
LLaVA (Sarcastic)  & 17.19 & 00.28 & /     & /     & /     & 13.34 & 00.39 & /     & /     & /     \\ 
LLaVA (CofiPara)  & 32.92 & 50.47 &  12.26    &  32.21    &  10.11    & 33.46 & 50.91 &  12.84    & 31.29     & 10.88     \\ \hline
Qwen-VL (Zero-shot)                     & 18.98 & 00.49 & 00.30 & 01.32 & 00.08 & 14.82 & 01.35 & 00.32 & 01.43 & 00.06 \\ 
Qwen-VL (CoT)                 & 18.70 & 00.49 & 00.68 & 02.57 & 00.13 & 15.09 & 01.48 & 00.62 & 02.65 & 00.07  \\
Qwen-VL (Sarcastic)           & 18.99  & 00.49 & 00.57 & 02.58 & 00.13 & 14.69 & 01.49 & 00.63 & 02.37 & 00.07\\
Qwen-VL (CofiPara)                    & {31.96} & {49.53} & {15.38} & {34.29} & {15.57} & {32.26} & {50.27} & {13.79} & {32.49} & {12.01} \\ \toprule
\end{tabular}}
\vspace{-0.3cm}
\caption{Effect of different LMMs on MSTI2.0 dataset.}
\label{lmm_res4msti}
\vspace{-0.4cm}
\end{table*}

Table~\ref{lmm_res4msti} shows the effect of different LMMs (i.e., LLaVA~\citep{liu2023visual} and Qwen-VL~\citep{bai2023qwen}) on the goal task MSTI, to enhance the comprehensiveness and robustness of the evaluation with different prompting strategies: 1) Zero-shot: Directly prompt a representative LMM, to identify sarcasm targets; 2) CoT: Prompt the LMMs with the Chain-of-Thought reasoning; 3) Sarcastic: Prompt the LMMs with the generated rationales by itself from the sarcastic position; 4) CofiPara: Our proposed paradigm under full setting based on the integration of the reasoning knowledge from LMMs and the pre-training knowledge from MSD, where LMMs are LLaVA or Qwen-VL. Note that LLaVA only supports to perform textual sarcasm target identification with multimodal input. We can see from the evaluations of broader LMM-based models like LLaVA and Qwen-VL that, the direct deployment based on LMMs in the zero-shot, CoT or Sarcastic settings, struggles without lightweight design specific to the MSTI task, also implying the importance of divergent thinking with LMMs to alleviate the inherent bias and variation during deductive reasoning for multimodal sarcasm target identification.

\section{Effect of LMM-generated Rationale}
As this work is the first to introduce LMMs into the multimodal sarcasm research area, all the MSD and MSTI baselines are not LMM-based. Although incidentally achieving such a more competitive performance on the MSD task as a by-product, our coarse-to-fine paradigm essentially focuses on the MSTI task.  Please note that our proposed divergent thinking with LMM is orthogonal to the conventional MSD approaches. Although integration of our proposed LMMs' divergent thinking with the conventional MSD methods may lead to further performance improvements, it is not our research focus in this work. As all the MSD baselines cannot be applied directly to the goal task (i.e., multimodal sarcasm target identification), simply adding the LMM-generated rationales as input of the MSD baselines does not have any practical significance towards the goal task MSTI.

In contrast, we argue that it is necessary to augment the representative MSTI baselines with LMM-generated rationales for a fair comparison with our proposed framework. We first present Table~\ref{rationale_msti} to illustrate how the two representative multimodal MSTI baseline models (i.e., Grounding DINO and MSTI-CB) would perform if equipped with LMM-generated rationales. One straightforward and possible solution would be: using the rationale generated from LMMs as input for an end-to-end training manner. Consequently, we use the rationale generated from the sarcastic perspective as the additional input for the baselines, similar to our paradigm in the finer-grained fine-tuning stage. We can see that both of the baselines cannot perform well after being equipped with the reasoning knowledge from LMMs. By this design, there may be two inherent weaknesses: 1) The knowledge elicited directly from LMMs may exhibit variation and bias; 2) If without our proposed divergent thinking mechanism with pre-training on MSD, the quality of generated rationales from LMMs needs some additional designs to filter. This reaffirms the advantage and necessity of our proposed coarse-to-fine paradigm for multimodal sarcasm target identification with LMMs.

\begin{table*}[] 
\centering
\resizebox{\textwidth}{!}{
\begin{tabular}{lcccccccccc}
\toprule
\multicolumn{1}{c}{}     & \multicolumn{5}{c}{Dev}               & \multicolumn{5}{c}{Test}              \\ \cmidrule(lr){2-6} \cmidrule(l){7-11} 
                         & EM    & F1    & AP    & AP50  & AP75  & EM    & F1    & AP    & AP50  & AP75  \\ \hline
Grounding DINO (LLaVA)   & 18.41 & 00.32 & 06.82 & 10.69 & 06.47 & 14.82 & 00.44 & 05.28 & 09.89 & 04.76 \\
Grounding DINO (Qwen-VL) & 18.15 & 00.01 & 05.20 & 08.93 & 04.18 & 14.55 & 00.44 & 05.23 & 10.69 & 04.87 \\
MSTI-CB (LLaVA)          & 27.78 & 44.31 & 02.57 & 08.53 & 00.05 & 29.24 & 48.15 & 01.55 & 06.24 & 00.05 \\
MSTI-CB (Qwen-VL)        & 28.61 & 44.81 & 00.89 & 04.49 & 00.02 & 29.11 & 47.32 & 00.86 & 05.05 & 00.01 \\
CofiPara-\textsc{Msti} (LLaVA)    & 32.92 & 50.47 & 12.26 & 32.21 & 10.11 & 33.46 & 50.91 & 12.84 & 31.29 & 10.88 \\
CofiPara-\textsc{Msti} (Qwen-VL)  & 31.96 & 49.53 & 15.38 & 34.29 & 13.39 & 32.26 & 50.27 & 13.79 & 32.49 & 12.01 \\ \toprule
\end{tabular}}
\caption{The impact of rationales generated by different LMMs on different MSTI baselines.}
\label{rationale_msti}
\end{table*}

\begin{table*}[]
\resizebox{\textwidth}{!}{
\begin{tabular}{lcccccccccc}
\toprule
\multicolumn{1}{c}{}     & \multicolumn{5}{c}{Dev}               & \multicolumn{5}{c}{Test}              \\ \cmidrule(lr){2-6} \cmidrule(l){7-11} 
                         & EM    & F1    & AP    & AP50  & AP75  & EM    & F1    & AP    & AP50  & AP75  \\ \hline
CofiPara-\textsc{Msti} & 31.96 & 49.53 & 15.38 & 34.29 & 13.39 & 32.26 & 50.27 & 13.79 & 32.49 & 12.01 \\
w/o RF               & 28.80 & 45.75 & 09.10 & 22.15 & 06.18 & 30.24 & 49.59 & 08.82 & 19.75 & 07.25 \\
w/o RP                & 28.80 & 49.54 & 11.09 & 25.98 & 08.27 & 31.72 & 50.69 & 10.14 & 21.86 & 09.15 \\
w/o LMM               & 31.13 & 47.30 & 08.87 & 21.09 & 06.76 & 30.91 & 48.32 & 07.50 & 19.36 & 04.22 \\ \toprule
\end{tabular}}
\caption{The influence of rationales at different training stages.}
\label{rationale_confipara}
\end{table*}

Then we delve into the effect of rationales in our different training stages as shown in Table~\ref{rationale_confipara}: 1) \textit{CofiPara-\textsc{Msti}} denotes using rationales generated from Qwen-VL at both training stages of coarser-grained pre-training and finer-grained fine-tuning, where pre-training uses the competing rationales and fine-tuning only uses the rationale from the sarcastic position; 2) \textit{w/o RF} denotes using competing rationales at pre-training and no rationale at fine-tuning; 3) \textit{w/o RP} denotes using the sarcastic rationale at fine-tuning without rationales as input at pre-training; 4) \textit{w/o LMM} denotes using no LMM-generated rationale at both stages. It can be demonstrated that the variants after removing rationales suffer different degrees of performance degradation, indicating the effectiveness of our design philosophy for multimodal sarcasm target identification by divergent thinking with LMM and multimodal fusion with small LM.

\section{Detailed Analysis of Competing Rationales}
To further analyze how the competing rationales applied in the pre-training stage affect the model training and decision-making process, we conduct more detailed and comprehensive ablation studies\footnote{Note that as the relative positions between the competing rationales are set as invariant, our preliminary verification indicated that the order information has minimal impact on the model's learning process. Therefore, here we do not delve further into this aspect.} on the competing rationales derived from Qwen-VL due to its best-performed results, as shown in Table \ref{mmsd_only_pos_neg} and Table \ref{msti_only_pos_neg}. In the results of coarser-grained pre-training in Table \ref{mmsd_only_pos_neg}, \textit{w/o $r_{neg}$} refers to only using the rationale $r_{pos}$ that explains why the sample is sarcastic as part of the input in the pre-training stage, likewise, \textit{w/o $r_{pos}$} indicates only using the rationale $r_{neg}$ that explains why the sample is non-sarcastic as part of the input in the pre-training stage. For finer-grained fine-tuning results in Table \ref{msti_only_pos_neg}, due to the nature of the MSTI task as emphasized in \S\ref{MSTI}, only the same rationale $r_{pos}$ from the sarcastic position can be used as part of the input in the fine-tuning stage, where \textit{w/o $r_{neg}$} and \textit{w/o $r_{pos}$} denote that the model parameters for fine-tuning are initialized from the models pre-trained on corresponding settings in Table \ref{mmsd_only_pos_neg}, respectively. 

From Table \ref{mmsd_only_pos_neg}, we observe that compared to CofiPara-\textsc{Msd} that uses competing rationales, both \textit{w/o $r_{neg}$} and \textit{w/o $r_{pos}$} showed sharp decreases in sarcasm detection, suggesting that introducing either $r_{pos}$ or $r_{neg}$ alone could bring severe noises that are one-way grossly misleading. The negative impact of adding $r_{neg}$ is larger than $r_{pos}$, as $r_{pos}$ relatively provides more information, which is instinctively understandable. In the fine-tuning stage, as shown in Table \ref{msti_only_pos_neg}, by removing different rationales at the pre-training stage, the models showed performance drops to varying extents. The fact that \textit{CofiPara-\textsc{Msti}} has the best performance compared to other settings proves that competing rationales in pre-training indeed make our model more robust to noisy LMM rationales, as well as facilitating reasoning. On the other hand, \textit{w/o $r_{pos}$} and \textit{w/o $r_{neg}$} settings both outperform \textit{w/o RP} that uses no rationales in pre-training, showing that using either rationale could bring a certain level of improvement in robustness, which accords with our naive idea. Moreover, the competing rationales could further encourage divergent thinking for multimodal sarcasm by resorting to a fundamental characteristic of human problem-solving, i.e., competing statements.
Overall, the comprehensive results reaffirm that our proposed competing rationales for coarser-grained pre-training could better contribute to multimodal sarcasm target identification.

\begin{table}[] 
\centering
\resizebox{0.9\linewidth}{!}{\begin{tabular}{lcccc}
\toprule
\multicolumn{1}{c}{}       & \multicolumn{1}{l}{Acc.} & P     & R     & F1    \\ \hline
CofiPara-\textsc{Msd} & 85.70  & 85.96 & 85.55 & 85.89 \\ 
w/o LMM & 81.61                    & 81.92 & 81.36 & 81.46 \\
w/o $r_{neg}$ & 68.55  & 68.68 & 68.49 & 68.99 \\ 
w/o $r_{pos}$ & 58.57  & 51.94 & 53.28 & 49.10 \\ 
\toprule 
\end{tabular}}
\caption{The influence of competing rationales on multimodal sarcasm detection.}
\label{mmsd_only_pos_neg}
\end{table}

\begin{table*}[]
\resizebox{\textwidth}{!}{
\begin{tabular}{lcccccccccc}
\toprule
\multicolumn{1}{c}{}     & \multicolumn{5}{c}{Dev}               & \multicolumn{5}{c}{Test}              \\ \cmidrule(lr){2-6} \cmidrule(l){7-11} 
                         & EM    & F1    & AP    & AP50  & AP75  & EM    & F1    & AP    & AP50  & AP75  \\ \hline
CofiPara-\textsc{Msti}        & 31.96 & 49.53 & 15.38 & 34.29 & 13.39 & 32.26 & 50.27 & 13.79 & 32.49 & 12.01 \\
w/o RP                & 28.80 & 49.54 & 11.09 & 25.98 & 08.27 & 31.72 & 50.69 & 10.14 & 21.86 & 09.15 \\
w/o $r_{neg}$           & 31.82 & 49.19 & 12.57 & 26.23 & 12.32 & 31.85 & 49.41 & 12.63 & 28.15 & 10.94 \\
w/o $r_{pos}$           & 31.13 & 49.54 & 14.99 & 31.85 & 11.72 & 29.97 & 50.39 & 13.11 & 26.45 & 10.43 \\
\toprule
\end{tabular}}
\caption{The influence of competing rationales on multimodal sarcasm target identification. \textit{w/o $r_{neg}$} and \textit{w/o $r_{pos}$} mean that the model parameters for fine-tuning are initialized from the pre-training stage under different settings, where only the rationale from sarcastic or non-sarcastic perspectives is used as part of the input, respectively, instead of our proposed competing rationales with both sarcastic and non-sarcastic positions.}
\label{msti_only_pos_neg}
\end{table*}

\section{Effect of Different Proposed Components}
Besides the core ablative test results on the MSTI2.0 test set shown in Table~\ref{abla_msti_test} and the supplemental results on the dev set shown in Table~\ref{abla_msti_dev}, we further perform ablation studies by discarding some important components of our model: 1) \textit{w/o QSM}: Use global visual features instead of query-selected local features for cross-modality text decoding.
2) \textit{w/o CTD}: Remove the image-to-text cross-attention in the text decoder, and only use textual features for text decoding. 
3) \textit{w/o CID}: Remove the image-to-text and text-to-image cross-attention in the image decoder, and only use self-attention in image decoding.

As demonstrated in Table~\ref{QSM}, it can observed that, without using the query-selection mechanism, our model showed a slight improvement in EM and F1 score, while AP50 downgraded by 2.09\% and 1.20\%. This can be explained as, by using query-selected local features, our model indeed missed information in
identifying text targets. However, the reason why query selection benefits visual performance, we speculate, is that by fusing local features that are more text-related, the text encoder of our model can learn representations that correlate more to the key components in the images, thus benefiting the follow-up image decoding process. The performance degradation in \textit{w/o CTD} can be elucidated by similar reasons, the existence of cross-modality text decoder forced text encoder to learn representations that benefit cross-modality fusion in images, further facilitating image decoding. For \textit{w/o CID} setting, we observed that with text-to-image and image-to-text cross-attention removed, there's a drastic drop in visual target identification performance, indicating that the model loses its object detection ability, which proves that cross-modality fusion in image decoding is essential.

\begin{table*}[]
\resizebox{\textwidth}{!}{
\begin{tabular}{lcccccccccc}
\toprule
\multicolumn{1}{c}{}     & \multicolumn{5}{c}{Dev}               & \multicolumn{5}{c}{Test}              \\ \cmidrule(lr){2-6} \cmidrule(l){7-11} 
                         & EM    & F1    & AP    & AP50  & AP75  & EM    & F1    & AP    & AP50  & AP75  \\ \hline
CofiPara-\textsc{Msti}        & 31.96 & 49.53 & 15.38 & 34.29 & 13.39 & 32.26 & 50.27 & 13.79 & 32.49 & 12.01 \\
w/o QSM              & 32.92 & 50.47 & 12.26 & 32.21 & 10.11 & 33.46 & 50.91 & 12.84 & 31.29 & 10.88 \\
w/o CTD              & 30.58 & 49.35 & 12.91 & 26.48 & 11.31 & 32.25 & 51.87 & 11.96 & 25.44 & 11.34 \\
w/o CID              & 31.82 & 49.87 & 00.18 & 00.83 & 00.02 & 31.72 & 51.84 & 00.38 & 01.76 & 00.04 \\ \toprule
\end{tabular}}
\caption{The influence of query-selection mechanism (\textit{w/o QSM}), cross-modality text decoder (\textit{w/o CTD}), and cross-modality image decoder (\textit{w/o CID}).}
\label{QSM}
\end{table*}

\begin{table}[]
\resizebox{\linewidth}{!}{\begin{tabular}{lccccc}
\toprule
                               & EM    & F1    & AP    & AP50  & AP75  \\ \hline
CofiPara-\textsc{Msti}                  & 31.96 & 49.53 & 15.38 & 34.29 & 15.57 \\
w/o MSD               & 31.41 & 48.78 & 13.15 & 28.28 & 12.14 \\
w/o LMM                 & 31.13 & 47.30 & 08.87 & 21.09 & 06.76 \\
w/o MSD\&LMM & 29.21 & 48.51 & 08.82 & 23.92 & 05.07 \\ \toprule
\end{tabular}}
\caption{{Ablation results on MSTI2.0 dev set. Results of the test set are provided in Table \ref{abla_msti_test}.}}
\label{abla_msti_dev}
\end{table}

\section{Evaluation of Rationale Quality}
\label{erq}

\begin{table}[]
\resizebox{\linewidth}{!}{\begin{tabular}{lcccc}
\toprule
                & \multicolumn{2}{c}{GPT-4V} & \multicolumn{2}{c}{Human} \\ \cmidrule(lr){2-3} \cmidrule(l){4-5} 
                & LLaVA      & Qwen-VL      & LLaVA      & Qwen-VL      \\ \hline
Conciseness     & 2.00       & 2.52         & 2.29       & 2.31         \\
Informativeness & 2.08       & 2.38         & 2.45       & 2.60         \\
Persuasiveness  & 2.08       & 2.34         & 2.02       & 2.22         \\
Readability     & 2.77       & 2.54         & 2.69       & 2.69         \\
Soundness       & 2.27       & 2.23         & 2.21       & 2.35         \\ \toprule
\end{tabular}}
\caption{Automatic GPT-4V and human evaluation of LMM-generated rationales on MSTI2.0 dataset.}
\label{gpt4_val}
\end{table}

\textbf{Automatic Evaluation.} Generally, there is no gold explanation about multimodal sarcasm for the multimodal sarcasm target identification task due to the diverse forms of textual expression. Devising reliable metrics without reference is not a straightforward task and can also be problematic. Furthermore, different types of text necessitate the evaluation of distinct aspects, such as informativeness, fluency, soundness, etc.~\cite{fabbri2021summeval, mehri2020unsupervised}, which makes it hard to design metrics for each type of text and dimension separately. Nowadays, GPT-4V~\cite{OpenAI2023GPT4TR} has revolutionized the field of LMMs with a more powerful expressive capacity for multimodal inputs. In this subsection, we present a new automatic evaluation using GPT-4V in a reference-free mode, to evaluate the text quality of the explanations generated by our approach from LLaVA and Qwen-VL. 

We randomly selected 50 samples from the MSTI2.0 test set.
Specifically, GPT-4V is prompted to score the explanations w.r.t. each multimodal sample according to the following criteria: 1) \textit{Conciseness}: the explanation contains less redundant information; 2) \textit{Informativeness}: the explanation provides new information, such as explaining the background and additional context; 3) \textit{Persuasiveness}: the explanation seems convincing; 4) \textit{Readability}: the explanation follows proper grammar and structural rules; 5) \textit{Soundness}: the explanation seems valid and logical. For each criterion, a 3-point Likert scale was employed, where 1 meant the poorest quality and 3 the best.

Table~\ref{gpt4_val} demonstrates the averaged scores of the explanation evaluation by GPT-4V on the two sources (i.e., LLaVA and Qwen-VL) regarding the five criteria. 
We could observe that: 1) Qwen-VL scores higher than LLaVA on Conciseness, Informativeness and Persuasiveness scores, indicating that Qwen-VL is more effective in providing information, as well as providing convincing explanations. 2) LLaVA, however, excels in Readability and Soundness scores, which means that it has a better capability of reasoning. 3) We also notice that rationales generated by Qwen-VL are longer than those of LLaVA, with average lengths of 120.98 and 99.68, which further proves that Qwen-VL offers more evidence in its explanations.

\textbf{Human Evaluation.}
Considering that automatic evaluation cannot realistically measure the quality of 
the chosen explanations generated by LMMs, we further conduct the human subjects study to evaluate the overall quality of explainability. For the 50 samples randomly selected in the previous automatic evaluation phase, 10 professional linguistic annotators (five females and five males between the ages of 26 and 29) are asked to evaluate the explanations of our model from LLaVA and Qwen-VL. The metrics of human evaluation are the same as the \textit{automatic evaluation}. 

To protect our human evaluators, we establish three guidelines: 1) ensuring their acknowledgment of viewing potentially uncomfortable content, 2) limiting weekly evaluations and encouraging a lighter daily workload, and 3) advising them to stop if they feel overwhelmed. Finally, we regularly check in with evaluators to ensure their well-being.

The scores of human evaluation are shown in Table~\ref{gpt4_val}. Note that the intra-class agreement score is 0.651. The average Spearman’s correlation coefficient between any two annotators is 0.632. We can observe that: 1) Qwen-VL explanations are more human-preferable, beating LLaVA on almost every metric. 2) Qwen-VL mainly shows advantages in Informativeness, Persuasiveness and Soundness scores, denoting that its superiority lies in both reasoning and resolving sarcastic shreds of evidence. 3) According to our evaluators, Qwen-VL and LLaVA both show excellent fluency in the generated contents, where LLaVA tends to use clauses to make overall descriptions, while Qwen-VL favors short sentences that depict detailed facts. 4) We notice that human and automatic evaluations show deviation in Readability and Soundness. We conjecture GPT-4V prefers LLaVA over Qwen-VL for the reason that, LLaVA explicitly used GPT-4 generated instruction tuning data, which makes LLaVA rationales slightly more fluent and logical to the taste of GPT-4V. On the other hand, Qwen-VL applied a cleaning strategy on classical pre-training tasks, removing a large proportion of data. In addition, it added tasks like OCR and grounding, which may change its overall reasoning paths into patterns not so favorable to GPT-4V, but make sense from human perspectives.

\section{Error Analysis}
\begin{figure}[t]
    \centering
    \includegraphics[width=1.\linewidth, scale=1.00]{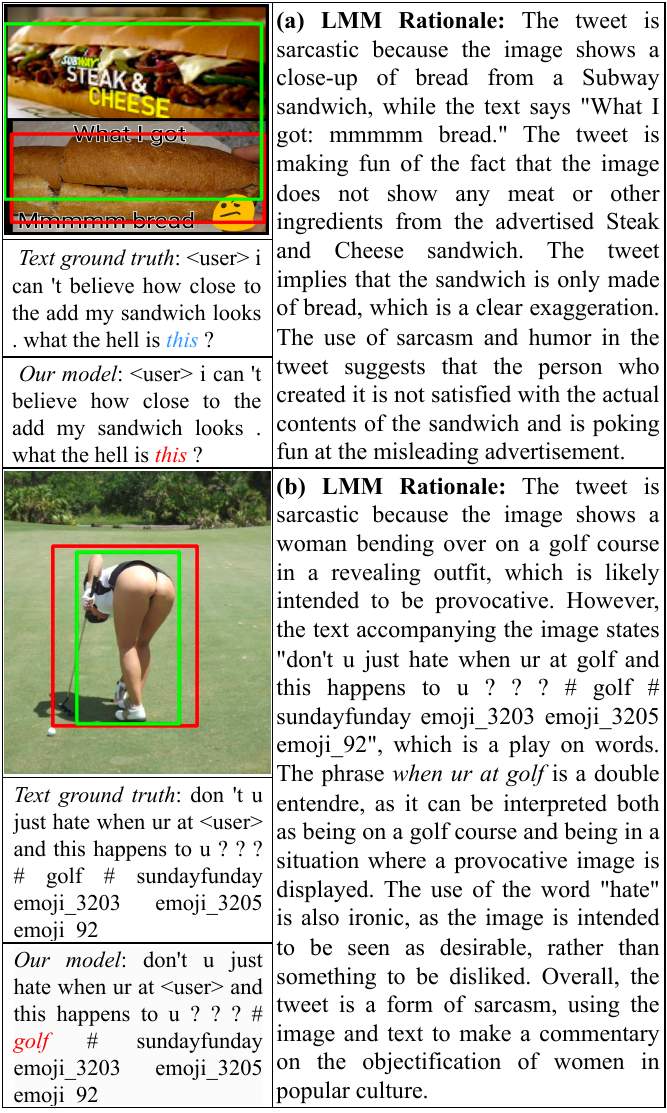}
    \caption{Examples of wrongly identified samples by our proposed CofiPara framework for MSTI.} 
    \label{error_analysis}
\end{figure}

Apart from the case study in \S\ref{case_study_sec}, we also conduct an error analysis for a better understanding of our proposed method. Figure \ref{error_analysis} shows two of the incorrectly predicted examples\footnote{\color{red}\textbf{Disclaimer:} This section contains content that may be disturbing to some readers.}, where the truth boxes in the image are labeled in red rectangles and the predicted boxes are the green ones. For text targets, we labeled our outputs in red italics, and ground truth in blue.

In the first example, the sarcasm in the original message refers to the poor sandwich that almost looks like two pieces of bread. While in LMM rationale, this tweet is over-analyzed as sarcastic towards SUBWAY’s false advertisement, which misleads our model to generate the box that covers both the ad and the sandwich. Although LMM makes a good explanation of this example, the information it provides is partly unnecessary and may lead to false predictions.

As for the second example, where our model identifies a correct visual target but a wrong textual target. In the explanation, LMM misinterpreted the phrase “golf” as a word for a provocative image. Although the explanation reached the correct answer in the end that the sarcasm lies on the contrary between the “desiring” situation and the use of the word “hate”, this situation is wrongly correlated to the word “golf”, misleading our model to give a false prediction.

From the error analysis, we can also find that multimodal sarcasm target identification is more challenging than unimodal sarcasm target identification because it requires the integration and interpretation of heterogeneous data sources, i.e., texts and images, to understand the nuanced and often contradictory nature of sarcasm. This complexity demands sophisticated models capable of capturing and correlating features across different modalities to accurately identify the target of sarcasm, which is inherently more difficult than analyzing a single type of data.

\section{More Examples of Explainability}
\begin{figure*}
\centering
\scalebox{0.90}{
\includegraphics[width=1\linewidth]{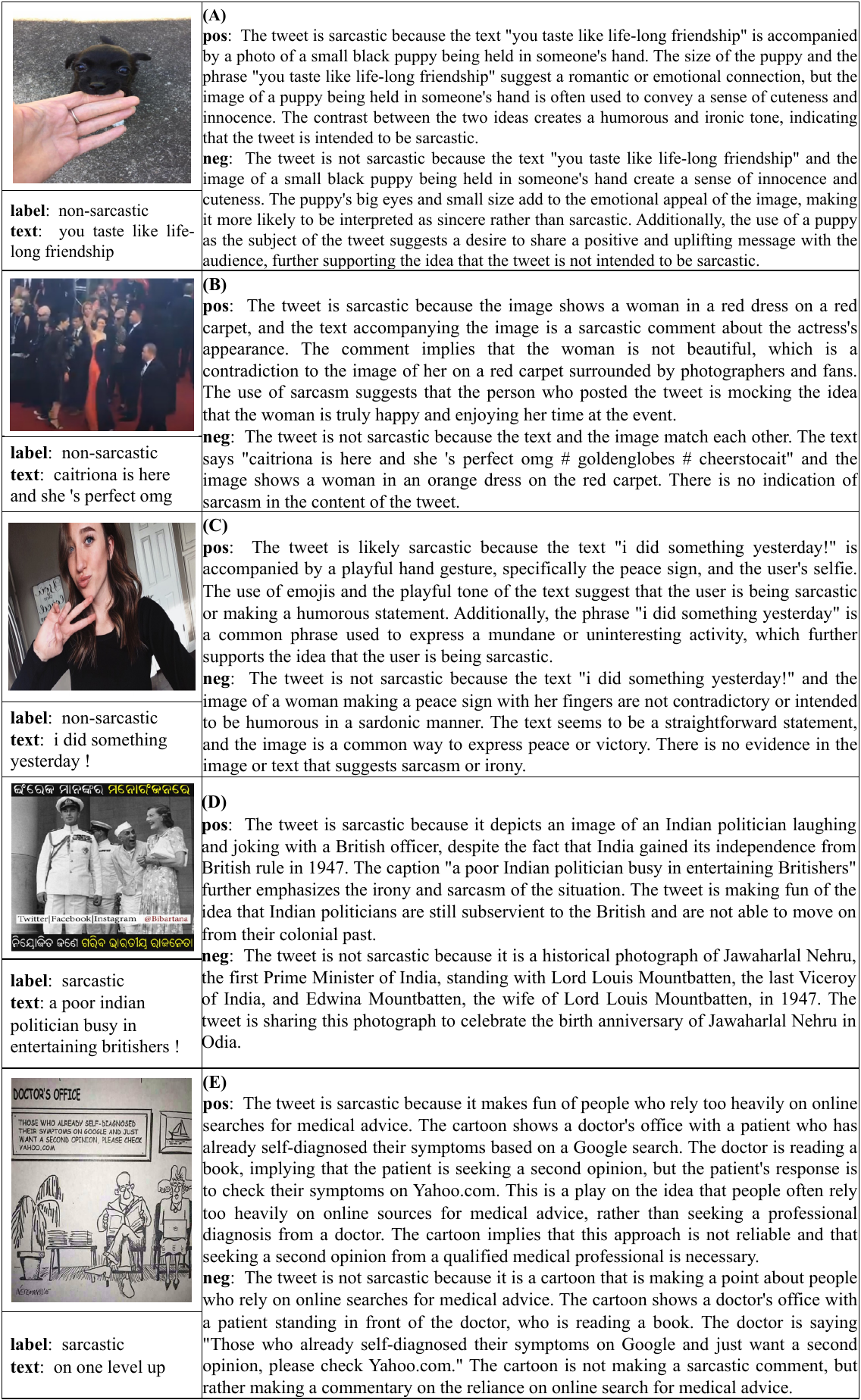}}
\caption{Examples of competing rationales at the coarser-grained pre-training stage of our proposed CofiPara.}
\label{example1}
\end{figure*}

\begin{figure*}
\centering
\scalebox{0.90}{
\includegraphics[width=1\linewidth]{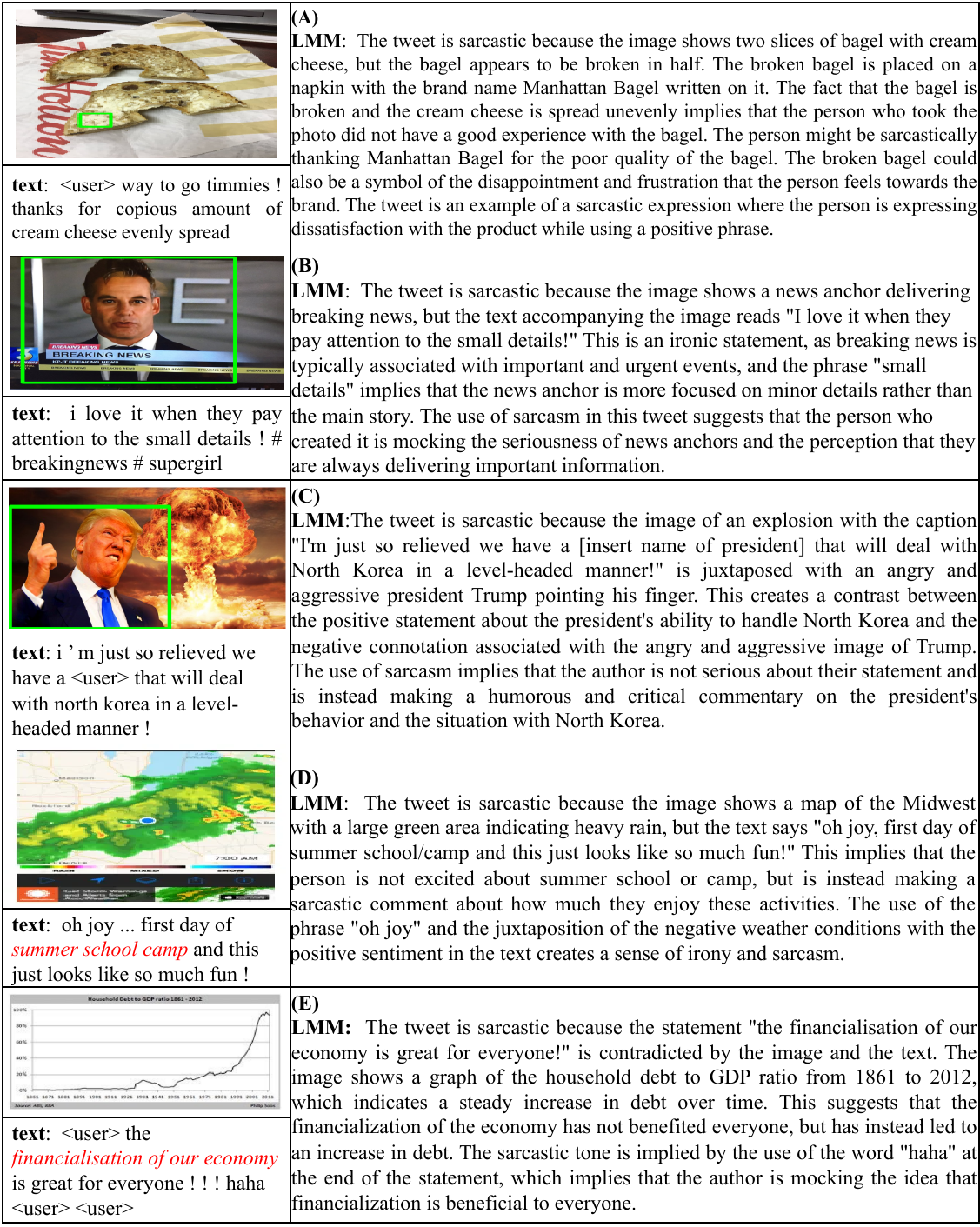}}
\caption{Examples of sarcasm explainability of our proposed CofiPara.}
\label{example2}
\end{figure*}

We present more examples in Figure~\ref{example1} and Figure~\ref{example2} to illustrate the supportive basis for the explanation of multimodal sarcasm.
From the explanations, it can be observed that the LMM usually explains sarcasm in a certain mode, by describing first and then reasoning. In its descriptions, LMM can extract key information like political figures, historical scenes and trends in graphics, and by combing texts and images, it can also decipher some of the desensitized words like \textit{<user>} that refer to a certain person in the context. We also noticed that, in its reasoning procedure, the LMM tends to perceive sarcasm by using the incongruity between texts and images, and the non-sarcastic samples are defined as non-contradictory, which is quite similar to human ways of thinking.

\section{Discussion about LMMs}
In this section, we discuss potential concerns on LMMs in the following four aspects: 1) \textbf{Reproducibility}: Since our proposed method uses the open-source LMMs like LLaVA and Qwen-VL as our backbone model instead of the close-source GPT-4V, the results are definitely reproducible with open-source codes. In order to make sure the LLaVA or Qwen-VL can generate the same contents for the same instance, we utilized the specific version `LLaVA-1.5 13B' of LLaVA and `v1.0.0' of Qwen-VL, and further set the parameter temperature as 0 without sampling mechanism, that is, the greedy decoding was adopted to ensure the deterministic results for the explanation generation with the same prompt. 2) \textbf{High-quality Explanations}: Although both LLaVA and Qwen-VL are well-known excellent LMMs, there still could be noise and the quality of reasoning from them may vary. This is indeed a tricky issue for us to consider in this work, for which we have proposed both the `Divergent Thinking with LMM' mechanism and the coarse-to-fine training paradigm to ensure the quality of reasoning utilized in our proposed model. The coarser-gained pre-training stage combined with the competing rationales derived from the divergent thinking of LMMs could facilitate our model to be more resilient and implicitly distill useful information in the following finer-grained fine-tuning stage. Furthermore, as illustrated in \S\ref{erq}, we also make a systematic evaluation to assess the quality of the rationale explainability from both automatic and human perspectives. 3) \textbf{Test Set Leakage}: The test set leakage issue does not exist in LLaVA or Qwen-VL as their papers have clearly described the instruction-tuning data used for training, which does not include any multimodal sarcasm dataset used in our task. As shown in Table~\ref{lmm_res} and Table~\ref{lmm_res4msti}, where the LLaVA or Qwen-VL was directly deployed to test on the two standard datasets, LLaVA or Qwen-VL did struggle on the datasets. When replacing LLaVA with Qwen-VL in our framework, we can consistently observe enhanced performance. This suggests that such improvement is basically attributed to our designed framework rather than test set leakage, since directly prompting LLaVA or Qwen-VL should result in fairly good performance if it takes advantage of test set leakage. 4) \textbf{Generalizability}: We believe our CofiPara framework is a general technique that works with LMMs, because our method does not choose GPT-4V as the representative LMM to generate rationales, but works well with the open-source LLaVA or Qwen-VL, which is not an OpenAI system.

\section{Dicussion about MSD and MSTI}
Generally, for multimodal sarcasm moderation, MSD is the initial step of determining the presence of sarcasm in an expression at the holistic semantics~\citep{qin2023mmsd2}, and MSTI goes a step further to analyze the specific focus of the sarcasm~\citep{joshi2018sarcasm}. As MSD should be conducted first to establish the presence of sarcasm at the coarse level, followed by MSTI to explicitly identify the specific targets of the detected sarcasm at the fine level~\citep{joshi2018sarcasm}, we devise a cohesive framework to operate on the coarse-to-fine training paradigm, aimed at pinpointing nuanced visual and textual targets of multimodal sarcasm, where multi-task learning is obviously not suitable due to their cascaded correlation.

\section{Future Work}
We will explore the following directions in the future:
\begin{itemize}
    \item Explainability evaluation: Although this study is at the forefront of using GPT-4V for the automated evaluation of rationale quality, the results exhibit slight discrepancies when compared to human subject studies. Furthermore, integrating GPT-4V into the rationale generation phase necessitates the exploration of even more advanced language models for evaluating the explanations produced by GPT-4V. Therefore, there is a need for more precise automatic evaluation methods for explanation quality. Simultaneously, conducting more extensive human subject studies with a broader pool of evaluators in a systematic manner would further validate our findings.
    \item LMM prompting: Our initial approach involved heuristically designing a single-turn prompt for LMMs to facilitate divergent thinking. However, we observed instances where the generated text overlooked key details, such as the semantic context of the sample. Moving forward, we plan to refine our prompting strategy to incorporate multi-turn or structured interactions~\citep{lin2023zero} with LMMs. This will enable more effective activation of commonsense reasoning related to vulnerable targets within sarcastic content, enhance visual feature extraction, and foster improved multimodal reasoning. 
    \item High-quality benchmark: We plan to expand our research by developing additional high-quality MSTI benchmarks from perspectives like dataset construction on social media with propagation structure~\citep{lin2021rumor, lin2022detect, yang2023wsdms}, and updating our framework to incorporate a wider range of more powerful open-source LMMs, as they become available. This will enable us to further explore and evaluate the efficacy of LMMs in enhancing sarcasm awareness and understanding.
\end{itemize}

\end{document}